\documentclass{article} 
\usepackage{iclr2027_conference,times}

\usepackage{amsmath,amsfonts,bm}

\def\eqref#1{equation~\ref{#1}}

\def\1{\bm{1}}

\DeclareMathAlphabet{\mathsfit}{\encodingdefault}{\sfdefault}{m}{sl}
\SetMathAlphabet{\mathsfit}{bold}{\encodingdefault}{\sfdefault}{bx}{n}

\usepackage[table]{xcolor}
\usepackage{hyperref}
\usepackage{url}
\usepackage{graphicx}
\usepackage{float}
\usepackage{wrapfig}
\usepackage{dsfont}
\usepackage{booktabs}
\usepackage{multirow}
\usepackage{amssymb}
\usepackage{arydshln}

\usepackage[capitalize]{cleveref}
\crefname{section}{Sec.}{Secs.}
\Crefname{section}{Section}{Sections}
\Crefname{table}{Table}{Tables}
\crefname{table}{Tab.}{Tabs.}

\AddToHook{env/figure/begin}{\setlength{\abovecaptionskip}{0pt}}
\AddToHook{env/figure*/begin}{\setlength{\abovecaptionskip}{0pt}}
\AddToHook{env/wrapfigure/begin}{\setlength{\abovecaptionskip}{0pt}}

\definecolor{tablebest}{RGB}{255,224,192}
\definecolor{tablesecond}{RGB}{255,255,153}

\title{DensiTok: Making Feed-Forward 3D Gaussian Splatting See More Views Than It Is Given}

\author{%
Minhyeok Lee$^{1}$\thanks{These authors contributed equally to this work.} \quad
Jungho Lee$^{2}$\footnotemark[1] \quad
Minseok Kang$^{1}$ \\
Heeseung Choi$^{3}$ \quad
Ig-Jae Kim$^{3}$ \quad
Sangyoun Lee$^{1}$\thanks{Corresponding author.} \\[0.5em]
$^{1}$Yonsei University \quad $^{2}$NAVER AI Lab \\
$^{3}$Korea Institute of Science and Technology (KIST) \\
\texttt{\{hydragon516,louis0503,syleee\}@yonsei.ac.kr} \\
\texttt{jungho.lee\_v@navercorp.com} \quad \texttt{\{hschoi,drjay\}@kist.re.kr}
}

\iclrfinalcopy 
\begin{document}

\maketitle
\lhead{}

\begin{figure}[H]
	\centering
	\includegraphics[width=\textwidth]{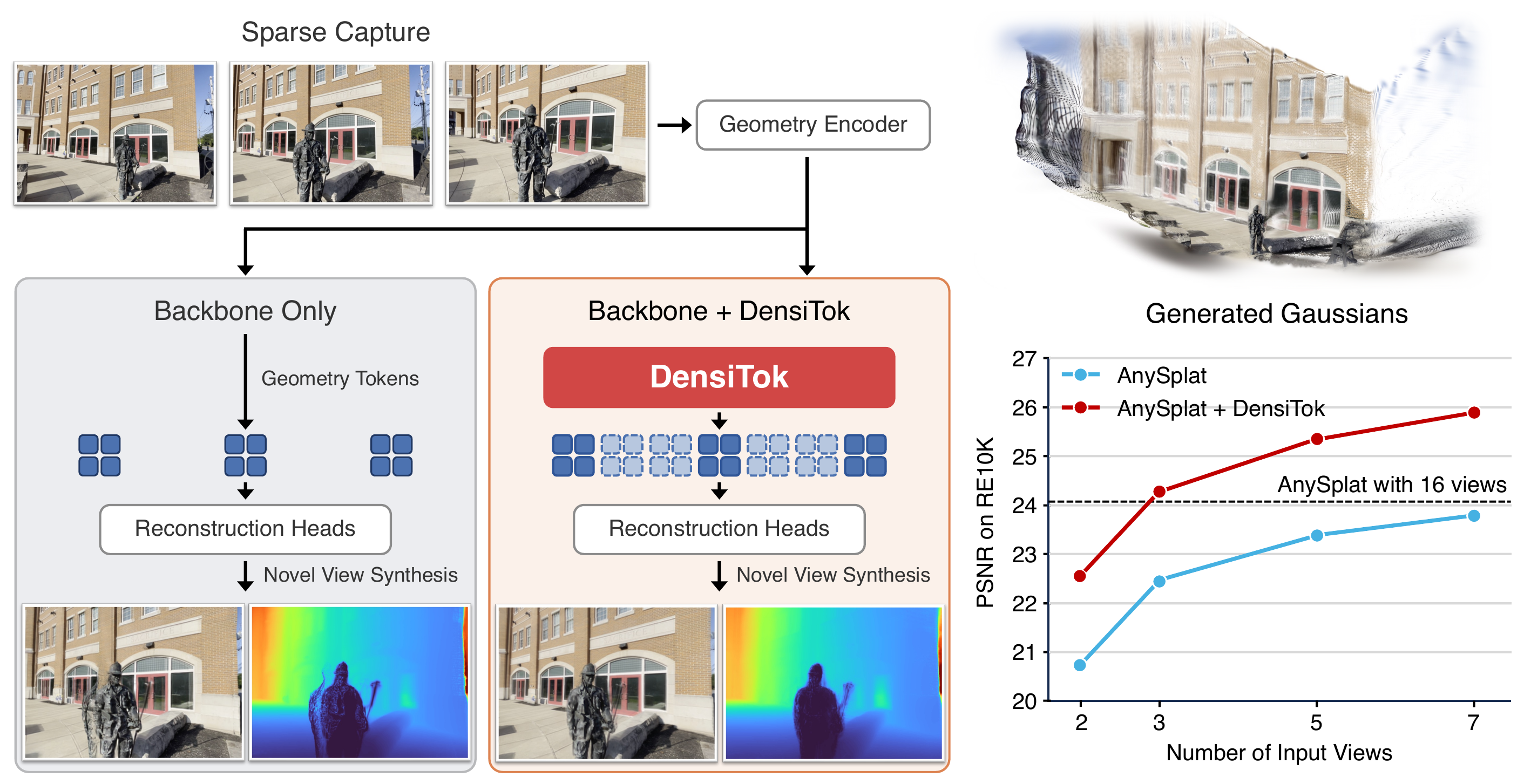}
	\caption{\textbf{DensiTok} augments geometry tokens for unobserved views while keeping the pretrained encoder and reconstruction heads frozen. With only three input views, it enables AnySplat to exceed its 16-view baseline in PSNR on RealEstate10K.}
	\label{fig:fig1_overview}
\end{figure}

\begin{abstract}
Feed-forward 3D Gaussian Splatting (3DGS) reconstructs a scene in a single forward pass, replacing per-scene optimization with a network trained across many scenes.
Its quality, however, degrades sharply as the number of input images drops.
The bottleneck is upstream of the reconstruction heads: from a few unposed views, the internal representation they read carries no evidence for unobserved regions, leaving holes, floaters, and blur.
The common remedy supplies that evidence as pixels, synthesizing extra views with an image or video generator and re-encoding them, which is costly and not 3D-consistent by construction.
\textit{We instead densify the evidence itself.}
We present \textbf{DensiTok}, a plug-in module for pretrained feed-forward 3DGS models that densifies their internal geometry tokens directly, making a frozen backbone behave as though it had observed many more views than it was given.
DensiTok compresses those tokens into a compact latent space, completes the latents of the unobserved viewpoints in a single flow-matching step conditioned on camera geometry, and decodes them back into tokens that the original reconstruction heads.
The same module design can be integrated into different pretrained predictors while keeping each backbone and its reconstruction heads frozen. Completion in a low-dimensional latent space requires no image synthesis or additional encoder passes.
Across three pretrained backbones and two benchmarks, DensiTok consistently improves sparse-view reconstruction and recovers much of the gap to dense-view reconstruction.
Project page: \url{https://hydragon.co.kr/DensiTok}.
\end{abstract}

\section{Introduction}

Feed-forward 3D Gaussian Splatting predicts cameras, geometry, and appearance in a single network pass, replacing per-scene optimization with a model amortized over many scenes.
This makes reconstruction practical from unposed, unconstrained image collections, and recent backbones accept an arbitrary number of input views.
Such models~\citep{jiang2025anysplat,liu2025worldmirror,lin2025depth} share a common anatomy: a geometry encoder maps the input images to a set of view-aligned geometry tokens, and camera, depth, and Gaussian heads decode that shared representation into the final reconstruction.
Their quality, however, tracks the visual evidence they are given.
The right side of~\cref{fig:fig1_overview} plots reconstruction quality against the number of input views for several recent backbones, and it degrades in the few-view regime that motivated feed-forward reconstruction in the first place.
The cause sits upstream of the heads: a handful of unposed images gives the encoder little cross-view evidence, so its tokens describe only the surfaces that were actually observed.
Occluded and uncovered surfaces have no tokens at all, and the heads turn that gap into holes, floaters, and blurred novel views.
\textit{What the model lacks, then, is not pixels but tokens.}
This suggests a different target: rather than acquiring the views it never saw, synthesize the geometry tokens those views would have produced.

Existing remedies differ mainly in where they supply the missing evidence.
The dominant line of work operates in image space, either synthesizing extra input views with a multi-view or video diffusion model~\citep{gao2024cat3d, zhong2025taming}, or repairing those renderings, and refitting the scene to them~\citep{chen2024mvsplat360, yin2025gsfixer}.
Both incur two separate costs: running a large generative model at full image resolution, and then running the geometry encoder again over the enlarged set of views.
Because pixels are not a shared 3D representation, consistency across generated views is encouraged but never enforced, so the reconstruction must fit views that disagree with one another.
More fundamentally, pixels are a detour.
\textit{The reconstruction needs only tokens for the missing viewpoints.}
These methods produce entire RGB frames instead, which the encoder then compresses into those tokens, so most of what the generator produced is discarded immediately.
A second line of work supplies no new evidence at all, and instead trains feed-forward architectures specialized for sparse views, extracting as much as possible from the correspondences the input views do provide~\citep{chen2024mvsplat, xu2025depthsplat,smart2024splatt3r}.
Such models remain bounded by what those views cover, and they are standalone architectures tied to a small, fixed number of views.
Closest to our setting, recent work~\citep{jang2026repurposing} generates directly in token space, but without a compressed latent it must run cascaded multi-step generators on features with over a thousand channels, so generation stays expensive.

We introduce \textbf{DensiTok}, a plug-in that densifies the geometry tokens of a pretrained feed-forward 3DGS model so that its heads reconstruct as if from far more views.
As shown in~\cref{fig:fig1_overview}, DensiTok sits between the geometry encoder and the reconstruction heads, with the encoder frozen and the heads either frozen or minimally adapted.
The encoder's output is naturally organized as a sequence of per-view token sets, one for each input image.
DensiTok augments token sets for viewpoints that were never observed, so the heads receive the representation of a dense capture even though only a few images were encoded.
Doing this directly on the tokens is costly: the heads read tokens from several layers of the encoder, and each layer has over a thousand channels in current backbones.
DensiTok instead treats the whole stack of layers as one object: it compresses them into a single compact latent per view, completes the unobserved views in that latent space, and expands the result back into tokens for every layer.

Two components realize this design.
The first is a variational autoencoder (VAE)~\citep{kingma2013auto} with \textbf{Level-Adaptive Modulation (LAM-VAE)}.
Its encoder predicts how much each feature level should contribute for each token, fuses the features accordingly, and compresses the fused tokens into compact latents.
This lets a single latent stand in for all layers without fixing in advance which layers matter, since that varies across backbones and even across scenes.
The second is a Diffusion Transformer (DiT)~\citep{peebles2023scalable} for \textbf{Geometry Latent Augmentation}: it extends the latent sequence with the unobserved views, generating their latents from the observed ones in a single flow-matching~\citep{lipman2022flow} step, conditioned on their cameras.
Since the input is unposed, these cameras are not given.
The backbone's own camera head predicts them for the observed views, and the cameras of the new views are interpolated along the path through those predictions, so the new views densify the captured trajectory rather than extrapolate beyond it.
The VAE decoder then expands the completed latent sequence back into tokens at every layer the heads expect, and the camera, depth, and Gaussian heads decode them.
Image-space methods supply the missing evidence indirectly, as pixels that the encoder must process again.
DensiTok produces the evidence itself, once and in a compact latent space, in exactly the form the heads already consume.

We evaluate DensiTok on RealEstate10K~\citep{zhou2018stereo} and DL3DV~\citep{ling2024dl3dv} with three representative feed-forward 3DGS backbones: AnySplat~\citep{jiang2025anysplat}, WorldMirror~\citep{liu2025worldmirror}, and Depth Anything 3~\citep{lin2025depth}.
DensiTok consistently improves novel-view rendering quality from sparse, unposed inputs.
This comes at a bounded cost: the encoder still processes only the sparse inputs, and generation is a single step in a compact latent space, so where DensiTok reaches dense-view quality, it does so in less time than dense-view reconstruction itself.

\section{Related Work}
\label{sec:related_work}

\subsection{Feed-Forward Gaussian Splatting}

Feed-forward 3DGS amortizes reconstruction across scenes by using a single network to predict Gaussian attributes, cameras, and depth from a set of images in one pass.
Posed methods build explicit correspondence machinery, with pixelSplat~\citep{charatan2024pixelsplat} using epipolar matching, MVSplat~\citep{chen2024mvsplat} constructing cost volumes, and DepthSplat~\citep{xu2025depthsplat} adding monocular depth priors.
Splatt3R~\citep{smart2024splatt3r}, NoPoSplat~\citep{ye2025no}, PF3plat~\citep{hong2025pf3plat}, and SelfSplat~\citep{kang2025selfsplat} remove the pose requirement for sparse unposed inputs.
More recent general backbones, including AnySplat~\citep{jiang2025anysplat}, WorldMirror~\citep{liu2025worldmirror}, and Depth Anything 3~\citep{lin2025depth}, accept an arbitrary number of unconstrained views and jointly predict cameras, geometry, and Gaussians, typically by pairing a large geometry transformer with lightweight decoding heads.
All of these models, however, are bounded by what their inputs show.
With only a few views, overlap is limited, pose and depth are ambiguous, so the geometry tokens describe only the visible parts of the scene.
Therefore, they render everything the tokens do not cover as holes, floaters, and blurred novel views.
DensiTok instead treats sparse-view reconstruction as a completion problem inside the representation, augmenting token sets for additional viewpoints.

\subsection{Generative Priors for 3D Reconstruction}

Image-space generative priors are typically incorporated in two ways.
The first augments the sparse input set with pose-controlled synthetic views before reconstruction, as in CAT3D~\citep{gao2024cat3d}, VistaDream~\citep{wang2025vistadream}, and Bolt3D~\citep{szymanowicz2025bolt3d}.
The second starts from a coarse 3DGS reconstruction, renders it along novel camera trajectories, and uses a video or multi-view generative model to refine or complete the rendered sequence, as in MVSplat360~\citep{chen2024mvsplat360}, FlowR~\citep{fischer2025flowr}, VideoScene~\citep{wang2025videoscene}, GSFixer~\citep{yin2025gsfixer}, and G4Splat~\citep{ni2026g4splat}.
Both can synthesize plausible content for unobserved regions, but the generative step operates in RGB or generic video latent spaces with no shared 3D representation to respect.
Consequently, inconsistencies across the generated views turn into geometric artifacts.
Both also incur two separate costs: a generative model runs for many sampling steps at image resolution, and the scene must then be reconstructed from the enlarged set of views.
DensiTok instead generates in the backbone's own token space, so the completed representation is decoded into one Gaussian scene by the same heads, with no image synthesis and no second reconstruction.

\subsection{Generation in 3D-Aware Latent Spaces}

A separate line avoids image synthesis altogether by generating in a latent space that decodes straight to 3D.
LN3Diff~\citep{lan2025ln3diff++} diffuses a 3D-structured latent decoded into neural fields, Prometheus~\citep{yang2025prometheus} generates multi-view RGB-D latents that a Gaussian decoder turns into pixel-aligned Gaussians, and Wonderland~\citep{liang2025wonderland} lifts camera-controlled video latents directly to 3DGS.
Most recently, GLD~\citep{jang2026repurposing} uses the features of a pretrained geometry encoder~\citep{wang2025vggt} as the latent space itself, applying flow matching to selected feature levels for feature-space novel-view synthesis and deriving the remaining levels by propagating the result through the frozen encoder.
This removes the need for a VAE, but the generated levels keep the encoder's full width of over a thousand channels. Therefore, GLD runs cascaded multi-step generators on them, decodes RGB with a separately trained decoder.
DensiTok differs on each point. It compresses all feature levels into a single four-channel latent and completes the sequence in one step to produce a Gaussian scene rather than images.

\section{Method}
\label{sec:method}

\subsection{Problem Formulation}
\label{sec:problem_formulation}

A feed-forward 3DGS model takes $N$ unposed images $I_1, \dots, I_N$ and reconstructs the scene in a single forward pass.
Internally, a geometry encoder $\mathcal{E}_{\mathrm{geo}}$, a deep transformer, processes all $N$ images jointly and represents each image as $K$ tokens, each a $d_f$-dimensional vector.
Such tokens exist after every layer of the encoder, but the pretrained reconstruction heads do not read only the last one: they read a fixed set of $L$ intermediate layers (e.g., 4 of the 24 layers), chosen when the backbone was trained.
We call these $L$ layers \textit{levels} and index them by $\ell = 1, \dots, L$ from shallow to deep.
The part of the encoder's output that the heads consume is therefore
\begin{equation}
	(\mathbf{F}_1, \dots, \mathbf{F}_L) = \mathcal{E}_{\mathrm{geo}}(I_1, \dots, I_N),
	\qquad
	\mathbf{F}_\ell \in \mathbb{R}^{N \times K \times d_f},
	\label{eq:encoder}
\end{equation}
where $\mathbf{F}_\ell$ collects the tokens of all $N$ views at level $\ell$.
The camera, depth, and Gaussian heads decode $(\mathbf{F}_1, \dots, \mathbf{F}_L)$ jointly into cameras, depth maps, and a Gaussian scene.
Everything the heads know about the scene is contained in these tokens.

In our setting, only $n~(< N)$ images are available.
We still reason about $N$ views in total: the $n$ observed views, for which we have images, and $(N - n)$ unobserved views, which are virtual viewpoints along the camera path that have no image.
We index all $N$ views by $i = 1, \dots, N$ and record which ones are observed with a binary mask $\mathbf{m} \in \{0,1\}^N$, with $m_i = 1$ for observed views and $m_i = 0$ for unobserved ones.
Running the encoder on the $n$ observed images gives tokens for those views only,
\begin{equation}
	(\mathbf{F}^{\mathrm{obs}}_1, \dots, \mathbf{F}^{\mathrm{obs}}_L) = \mathcal{E}_{\mathrm{geo}}(\{I_i \mid m_i = 1\}),
	\qquad
	\mathbf{F}^{\mathrm{obs}}_\ell \in \mathbb{R}^{n \times K \times d_f},
	\label{eq:encoder_obs}
\end{equation}
and the unobserved views have no tokens at all.
DensiTok fills this gap.
For every level $\ell$ it produces a full-size token tensor $\tilde{\mathbf{F}}_\ell \in \mathbb{R}^{N \times K \times d_f}$ that covers all $N$ views, generating the tokens of the $(N - n)$ unobserved views from the observed ones.
The target is what the encoder would have produced had all $N$ images been available, so that the heads can decode $(\tilde{\mathbf{F}}_1, \dots, \tilde{\mathbf{F}}_L)$ exactly as they decode $(\mathbf{F}_1, \dots, \mathbf{F}_L)$ in the full-view case.
We keep the geometry encoder and reconstruction heads frozen by default. However, due to the structure of feed-forward 3DGS reconstruction heads, we also consider minimal appearance adaptation, as detailed later.

\begin{figure}[t]
  \centering
  \includegraphics[width=\textwidth]{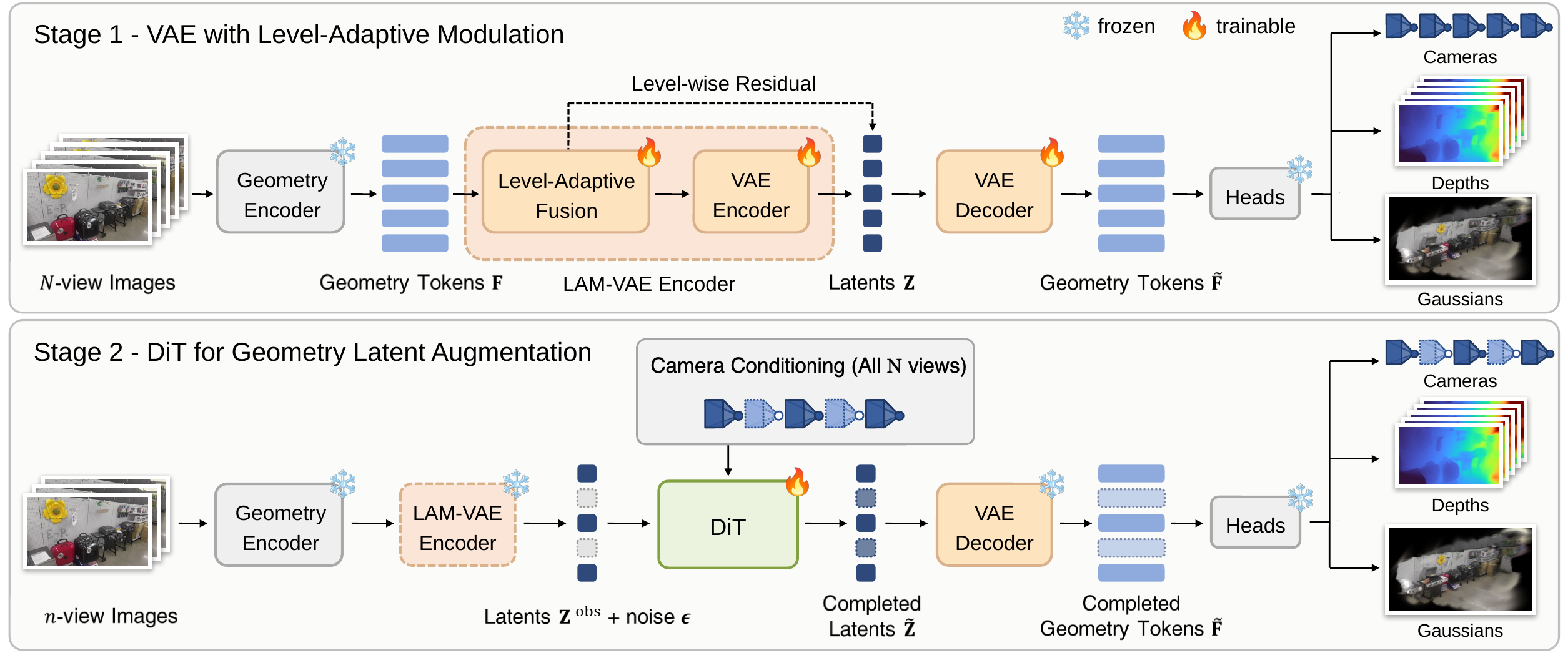}
  \caption{\textbf{Geometry token compression and completion.} Stage 1 trains a VAE with Level-Adaptive Modulation (LAM-VAE) to compress and reconstruct geometry tokens. level-adaptive fusion and the VAE encoder together form the LAM-VAE encoder, with a level-wise residual into the posterior mean. Stage 2 freezes the VAE and trains a camera-conditioned DiT to complete latents for unobserved views. The geometry encoder remains frozen throughout, while the reconstruction heads remain frozen or undergo minimal appearance adaptation.}
  \label{fig:lamvae}
\end{figure}

\subsection{VAE with Level-Adaptive Modulation}
\label{sec:level_adaptive_modulation}

As shown in~\cref{fig:lamvae}, Level-Adaptive Modulation (LAM-VAE) combines level-adaptive fusion with the VAE encoder.
We refer to this combined encoding module as the \textbf{LAM-VAE encoder}.

\paragraph{Level-adaptive fusion.}
Because the heads read all $L$ levels jointly, DensiTok has to produce $\tilde{\mathbf{F}}_\ell$ for every level, and generating each level with its own model would multiply the cost of an already high-dimensional problem by $L$.
We instead fuse the $L$ levels of every token into a single vector before compressing it into a $d_z$-dimensional latent, so that one VAE and one generator handle all $L$ levels.
Which levels carry the most useful information can vary across backbones and even across scenes, so rather than fixing a subset of levels in advance~\citep{jang2026repurposing}, we let each token decide how much every level contributes.
All operations in this paragraph act on one token at a time, so we drop the view and token indices and write $\mathbf{f}_\ell \in \mathbb{R}^{d_f}$ for the token's feature at level $\ell$.
Each $\mathbf{f}_\ell$ is first normalized with a LayerNorm~\citep{ba2016layer} specific to its level, so that no level dominates the others by magnitude.
A gating network then decides how much each level contributes.
It summarizes each $\mathbf{f}_\ell$ with a shared linear layer, concatenates the $L$ summaries, and outputs one logit per level through an MLP.
Because the MLP sees all $L$ summaries at once, the logit of each level depends on the others, so the gate compares levels rather than scoring each in isolation.
A softmax turns the logits into weights $\alpha_1, \dots, \alpha_L$ that sum to one.
We use gains $g_\ell = L\,\alpha_\ell$ rather than the weights themselves: the gains average to one, so uniform weights recover the plain concatenation of all levels, and gating acts as a reweighting around that baseline instead of shrinking the input by a factor of $L$.
The fused token is then a gain-weighted sum of level-wise linear projections,
\begin{equation}
	\mathbf{h} = \sum_{\ell=1}^{L} g_\ell \, \mathbf{W}_\ell \mathbf{f}_\ell,
	\label{eq:fusion}
\end{equation}
where $\mathbf{W}_\ell$ is a learnable linear projection that maps level $\ell$ to the width of the VAE encoder transformer.
Equivalently, one can concatenate the gain-weighted levels and apply a single linear layer.
The fused tokens of all $n$ observed views form one sequence that enters the VAE encoder transformer, which follows the alternating frame and global attention of VGGT~\citep{wang2025vggt}.
Frame blocks attend among the tokens of one view, global blocks attend across all views, and 2D rotary position embeddings~\citep{su2024roformer} encode each token's position on the patch grid.
For every token the transformer outputs a Gaussian posterior over a $d_z$-dimensional latent, parameterized by a mean $\boldsymbol{\mu}_0 \in \mathbb{R}^{d_z}$ and a log-variance $\log \boldsymbol{\sigma}^2 \in \mathbb{R}^{d_z}$, with $d_z = 4$ in our main setting.

\paragraph{Level-wise residual into the posterior mean.}
Compressing the fused token $\mathbf{h}$ into a $d_z$-dimensional latent, is what makes generation in a compact latent space possible.
However, such a narrow bottleneck can erase detail that is specific to a single level, especially since the levels have already been mixed in $\mathbf{h}$ before the transformer sees them.
Each normalized level feature $\mathbf{f}_\ell$ is projected into the latent space by a level-specific linear layer $\mathbf{P}_\ell \in \mathbb{R}^{d_z \times d_f}$, and the projections are combined with the same weights $\alpha_\ell$ from the fusion step,
\begin{equation}
	\boldsymbol{\mu} = \boldsymbol{\mu}_0 + \sum_{\ell=1}^{L} \alpha_\ell \, \mathbf{P}_\ell \mathbf{f}_\ell,
	\label{eq:residual}
\end{equation}
so the mean of each token is corrected by exactly the levels that dominate its fusion.
The residual touches only the mean; the log-variance $\log \boldsymbol{\sigma}^2$ predicted by the VAE encoder is kept as is.
During VAE training, the latent of each token is sampled as $\mathbf{z} = \boldsymbol{\mu} + \boldsymbol{\sigma} \odot \boldsymbol{\epsilon}_{\mathrm{vae}}$ with $\boldsymbol{\epsilon}_{\mathrm{vae}} \sim \mathcal{N}(\mathbf{0}, \mathbf{I})$.
At inference we drop the noise and use the mean itself as the deterministic latent representation, $\mathbf{z} = \boldsymbol{\mu}\in\mathbb{R}^{d_z}$.
Stacking the latents of all $K$ tokens of all encoded $N$ views gives $\mathbf{Z} \in \mathbb{R}^{N \times K \times d_z}$; in the completion setting only the $n$ observed views are encoded, yielding $\mathbf{Z}^{\mathrm{obs}} \in \mathbb{R}^{n \times K \times d_z}$, which is the input to the generator described next.
The LAM-VAE encoder includes this level-wise residual, so its complete path from geometry tokens to $\mathbf{Z}$ comprises level-adaptive fusion, the VAE encoder transformer, and the residual into the posterior mean.

\subsection{DiT for Geometry Latent Augmentation}
\label{sec:geometry_latent_augmentation}

We use a Diffusion Transformer (DiT)~\citep{peebles2023scalable} for geometry latent augmentation: given the observed latents $\mathbf{Z}^{\mathrm{obs}}$, it generates latents for the $(N - n)$ unobserved views and returns a completed tensor $\tilde{\mathbf{Z}} \in \mathbb{R}^{N \times K \times d_z}$ whose observed views are $\mathbf{Z}^{\mathrm{obs}}$.
The generated views are trained to match the corresponding views of $\mathbf{Z}$, the latents obtained when all $N$ images are encoded, which are available only at training time.
The DiT is conditioned on three inputs: the observed latents $\mathbf{Z}^{\mathrm{obs}}$; the mask $\mathbf{m}$ from the problem formulation~(\cref{sec:problem_formulation}), which marks the observed views; and cameras $\mathbf{g}_1, \dots, \mathbf{g}_N$ for all $N$ views, since an unobserved view has no image and its camera is the only information that specifies which view to generate.
How the cameras are obtained differs between training, where ground truth exists for all $N$ views, and inference, where only the cameras of the observed views can be predicted.

\paragraph{Latent sequence.}
$\mathbf{Z}^{\mathrm{obs}}$ covers only the $n$ observed views, whereas the DiT works on all $N$ views at once.
We therefore lay out an $N$-view sequence along the camera path.
The latents of each observed view sit at that view's position and stay clean throughout; they are the condition.
The positions of unobserved views are what the DiT fills in.
We use rectified flow matching~\citep{lipman2022flow,liu2022flow}, with $t = 0$ denoting clean latents and $t = 1$ pure noise.
During training, the latents of view $i$ at time $t$ are
\begin{equation}
	\mathbf{x}_{t,i} = m_i\, \mathbf{z}^{\mathrm{obs}}_i + (1 - m_i)\big[(1 - t)\, \mathbf{z}_i + t\, \boldsymbol{\epsilon}_i\big],
	\qquad \boldsymbol{\epsilon}_i \sim \mathcal{N}(\mathbf{0}, \mathbf{I}),
	\label{eq:flow_state}
\end{equation}
where $\mathbf{z}_i$ and $\mathbf{z}^{\mathrm{obs}}_i \in \mathbb{R}^{K \times d_z}$ are the latents of view $i$ in $\mathbf{Z}$ and $\mathbf{Z}^{\mathrm{obs}}$.
At inference, sampling starts from the $t = 1$, which requires only $\mathbf{Z}^{\mathrm{obs}}$ and noise.
Each view carries its own time $\tau_i = t(1 - m_i)$: observed views stay at the clean time zero, unobserved views follow the global time $t$.
This lets a single network handle both kinds of views in one sequence.

\paragraph{Camera conditioning and latent completion.}
Unobserved views enter as pure noise, so their cameras are the only signal that tells the DiT which view to generate.
We therefore encode the cameras of all $N$ views as Plücker rays.
For every patch token of every view, we compute the ray through that patch center, so the Plücker ray embedding has the same $K$-token layout as the latents.
The few tokens that do not correspond to a patch, such as the camera token, receive a learnable embedding instead.
These embeddings, together with the time $\tau_i$ (zero for observed views, $t$ for unobserved ones), condition every DiT block through adaptive LayerNorm (AdaLN)~\citep{peebles2023scalable}.
The blocks alternate frame and global attention in the same pattern as the VAE encoder, so that noisy unobserved views gather evidence from the clean observed views in the global blocks.
The DiT predicts the flow velocity with respect to increasing $t$, from clean latents toward noise, and the sampler integrates this velocity backward from $t = 1$ to $t = 0$ with Euler steps, resetting the observed views to $\mathbf{Z}^{\mathrm{obs}}$ after every step so they never drift; the result is the completed tensor $\tilde{\mathbf{Z}}$.
Training supervises only the unobserved views, including the finetuning that reduces sampling to a single Euler step.

\subsection{Camera Conditioning at Training and Inference}
\label{sec:camera_conditioning}

The DiT is conditioned on the cameras of all $N$ views, yet at inference only the $n$ observed views come with images, so the $N$ cameras have to be obtained differently at training and inference.
In both regimes, the backbone's pretrained camera head first predicts the cameras of the observed views from the observed tokens $\mathbf{F}^{\mathrm{obs}}$; we denote these predicted cameras by $\{\hat{\mathbf{g}}^{\mathrm{obs}}_i\}_{i=1}^n$. 

\paragraph{Training.}
During training, the dataset provides ground-truth cameras $\{\mathbf{g}^{\mathrm{gt}}_i\}_{i=1}^{N}$ for all $N$ views.
They cannot be used as $n$-view conditions directly: the backbone reconstructs at an arbitrary scale, so its predicted cameras $\{\hat{\mathbf{g}}^{\mathrm{obs}}_i\}_{i=1}^n$ live in a differently scaled world, and a DiT trained on raw ground-truth cameras would face a mismatch at inference, where only predicted cameras exist.
Both sets of cameras are expressed relative to the first observed view, following the convention of the backbone's camera head, so the two worlds already share their global rotation and translation and differ only in scale.
We remove this last difference by rescaling all ground-truth camera centers by a single factor, the ratio of the spread of the predicted centers to that of the ground-truth centers over the observed views, and denote the aligned cameras by $\{\mathbf{g}^{\mathrm{aligned}}_i\}_{i=1}^N$.
The DiT is trained on these aligned cameras and thus only ever sees cameras in the world it will encounter at inference.

\paragraph{Inference.}
At inference no ground-truth cameras exist, and the predicted cameras $\{\hat{\mathbf{g}}^{\mathrm{obs}}_i\}_{i=1}^n$ are themselves the reference, so no alignment is needed.
What is missing are the cameras of the $(N - n)$ unobserved views.
We obtain them by fitting a smooth trajectory through the predicted cameras, and sampling it at the unobserved views: camera centers are interpolated with a natural cubic spline, rotations with spherical interpolation.
We denote the resulting cameras by $\{\hat{\mathbf{g}}^{\mathrm{interp}}_i\}_{i=1}^{N-n}$ for $m_i = 0$.
Because the unobserved views are virtual, we place all of them between the first and last observed cameras, so the trajectory is only ever interpolated to densify the observed camera path.
In summary, the cameras that condition the DiT are the aligned ground-truth cameras $\{\mathbf{g}^{\mathrm{aligned}}_i\}_{i=1}^N$ during training, and at inference $\{\hat{\mathbf{g}}^{\mathrm{obs}}_i\}_{i=1}^n$ for observed views and $\{\hat{\mathbf{g}}^{\mathrm{interp}}_i\}_{i=1}^{N-n}$ for unobserved views.

\subsection{Overall Training Strategy}
\label{sec:overall_training_strategy}

DensiTok is trained in three stages, with the geometry encoder frozen throughout.
In Stage 1, the VAE with \textbf{Level-Adaptive Modulation} is trained to reconstruct the geometry tokens of the frozen geometry encoder.
In Stage 2, the DiT for \textbf{geometry latent augmentation} is trained on top of the frozen VAE with the rectified flow-matching objective on the unobserved views. At this stage, we can keep the Gaussian head frozen or minimally finetune it, as detailed in Appendix~\ref{sec:app_appfusion}.
In Stage 3, the model is finetuned for one-step sampling by fixing the time of the unobserved views at $t = 1$, so that a single Euler step maps noise directly to the completed latents.

\begin{table}[!t]
  \begin{center}
    \caption{Quantitative results on RealEstate10K and DL3DV with $n\in\{2,3,5,7\}$ sparse input views and dense-view references ($N=16$).}
    \resizebox{\columnwidth}{!}{%
      \centering
      \setlength{\tabcolsep}{2pt}
      \renewcommand{\arraystretch}{1.2}
      \begin{tabular}{l*{15}{c}}
        \toprule
        \multirow{3}{*}{Methods}
          & \multicolumn{7}{c}{RealEstate10K~\citep{zhou2018stereo}}
          & \multicolumn{7}{c}{DL3DV~\citep{ling2024dl3dv}} & \multirow{2}{*}{Runtime} \\
        \cmidrule(lr){2-8}\cmidrule(lr){9-15}
          & \multicolumn{3}{c}{Appearance} & \multicolumn{2}{c}{Camera} & \multicolumn{2}{c}{Depth}
          & \multicolumn{3}{c}{Appearance} & \multicolumn{2}{c}{Camera} & \multicolumn{2}{c}{Depth} & \\
        \cmidrule(lr){2-4}\cmidrule(lr){5-6}\cmidrule(lr){7-8}
        \cmidrule(lr){9-11}\cmidrule(lr){12-13}\cmidrule(lr){14-15}
          & PSNR$\uparrow$ & SSIM$\uparrow$ & LPIPS$\downarrow$ & AUC@$3^{\circ}\uparrow$ & AUC@$30^{\circ}\uparrow$ & $\delta_{1.25}\uparrow$ & AbsRel$\downarrow$
          & PSNR$\uparrow$ & SSIM$\uparrow$ & LPIPS$\downarrow$ & AUC@$3^{\circ}\uparrow$ & AUC@$30^{\circ}\uparrow$ & $\delta_{1.25}\uparrow$ & AbsRel$\downarrow$ & (s)$\downarrow$ \\
        \midrule
        \multicolumn{16}{l}{\emph{Dense-view} ($N=16$)} \\ [2pt]
        \quad AnySplat~\citep{jiang2025anysplat} & 24.04 & 0.792 & 0.189 & 20.4 & 77.5 & 0.835 & 0.151 & 22.30 & 0.683 & 0.246 & 24.1 & 85.4 & 0.803 & 0.158 & 0.434 \\
        \quad WorldMirror~\citep{liu2025worldmirror} & 25.11 & 0.856 & 0.161 & 41.8 & 83.3 & 0.880 & 0.110 & 24.47 & 0.819 & 0.173 & 69.0 & 94.0 & 0.825 & 0.141 & 0.495 \\
        \midrule
        \multicolumn{16}{l}{\emph{Sparse-view} ($n=2$)} \\ [2pt]
        \quad AnySplat & 20.73 & 0.706 & 0.250 & 20.0 & 77.1 & 0.828 & 0.168 & 19.85 & 0.626 & 0.287 & 22.5 & 85.8 & 0.774 & 0.186 & 0.100 \\
        \quad\textbf{+ DensiTok} & \textbf{22.56} & \textbf{0.766} & \textbf{0.223} & \textbf{26.7} & \textbf{78.9} & \textbf{0.846} & \textbf{0.153} & \textbf{22.40} & \textbf{0.699} & \textbf{0.255} & \textbf{30.4} & \textbf{88.9} & \textbf{0.801} & \textbf{0.170} & 0.258 \\ [2pt]
        \quad WorldMirror & 21.56 & 0.775 & 0.221 & 48.3 & 86.8 & 0.850 & 0.136 & 20.60 & 0.706 & 0.251 & 63.8 & 93.3 & 0.784 & 0.171 & 0.109 \\
        \quad\textbf{+ DensiTok} & \textbf{22.06} & \textbf{0.791} & \textbf{0.215} & \textbf{50.0} & \textbf{87.2} & \textbf{0.854} & \textbf{0.133} & \textbf{21.40} & \textbf{0.729} & \textbf{0.242} & \textbf{65.8} & \textbf{94.2} & \textbf{0.806} & \textbf{0.158} & 0.263 \\
        \midrule
        \multicolumn{16}{l}{\emph{Sparse-view} ($n=3$)} \\ [2pt]
        \quad AnySplat & 22.47 & 0.748 & 0.215 & 21.1 & 77.0 & 0.830 & 0.159 & 21.66 & 0.677 & 0.245 & 24.2 & 86.4 & 0.792 & 0.167 & 0.117 \\
        \quad\textbf{+ DensiTok} & \textbf{24.28} & \textbf{0.813} & \textbf{0.184} & \textbf{25.8} & \textbf{79.7} & \textbf{0.853} & \textbf{0.146} & \textbf{24.28} & \textbf{0.755} & \textbf{0.205} & \textbf{34.4} & \textbf{89.6} & \textbf{0.809} & \textbf{0.158} & 0.267 \\ [2pt]
        \quad WorldMirror & 23.29 & 0.816 & 0.190 & 48.3 & 86.4 & 0.868 & 0.119 & 22.54 & 0.768 & 0.201 & 66.7 & 94.3 & 0.810 & 0.152 & 0.133 \\
        \quad\textbf{+ DensiTok} & \textbf{23.81} & \textbf{0.836} & \textbf{0.181} & \textbf{49.6} & \textbf{87.0} & \textbf{0.874} & \textbf{0.117} & \textbf{23.35} & \textbf{0.787} & \textbf{0.189} & \textbf{70.0} & \textbf{95.5} & \textbf{0.823} & \textbf{0.147} & 0.277 \\
        \midrule
        \multicolumn{16}{l}{\emph{Sparse-view} ($n=5$)} \\ [2pt]
        \quad AnySplat & 23.39 & 0.768 & 0.203 & 20.6 & 76.6 & 0.832 & 0.155 & 22.58 & 0.690 & 0.234 & 24.3 & 85.5 & 0.799 & 0.161 & 0.160 \\
        \quad\textbf{+ DensiTok} & \textbf{25.35} & \textbf{0.838} & \textbf{0.162} & \textbf{25.2} & \textbf{78.8} & \textbf{0.852} & \textbf{0.141} & \textbf{25.33} & \textbf{0.782} & \textbf{0.182} & \textbf{33.5} & \textbf{87.7} & \textbf{0.816} & \textbf{0.153} & 0.299 \\ [2pt]
        \quad WorldMirror & 24.29 & 0.837 & 0.175 & 46.0 & 84.8 & 0.876 & 0.112 & 23.67 & 0.798 & 0.180 & 66.5 & 93.7 & 0.818 & 0.146 & 0.182 \\
        \quad\textbf{+ DensiTok} & \textbf{24.91} & \textbf{0.861} & \textbf{0.162} & \textbf{47.3} & \textbf{85.3} & \textbf{0.880} & \textbf{0.109} & \textbf{24.38} & \textbf{0.825} & \textbf{0.167} & \textbf{68.9} & \textbf{94.3} & \textbf{0.833} & \textbf{0.138} & 0.320 \\
        \midrule
        \multicolumn{16}{l}{\emph{Sparse-view} ($n=7$)} \\ [2pt]
        \quad AnySplat & 23.79 & 0.780 & 0.195 & 20.9 & 76.7 & 0.831 & 0.153 & 23.01 & 0.709 & 0.225 & 23.1 & 84.9 & 0.806 & 0.157 & 0.207 \\
        \quad\textbf{+ DensiTok} & \textbf{25.89} & \textbf{0.851} & \textbf{0.150} & \textbf{24.5} & \textbf{78.5} & \textbf{0.853} & \textbf{0.137} & \textbf{25.71} & \textbf{0.796} & \textbf{0.169} & \textbf{34.0} & \textbf{86.7} & \textbf{0.817} & \textbf{0.151} & 0.333 \\ [2pt]
        \quad WorldMirror & 24.75 & 0.847 & 0.167 & 44.5 & 84.1 & 0.880 & 0.111 & 24.28 & 0.816 & 0.170 & 63.7 & 92.7 & 0.824 & 0.143 & 0.235 \\
        \quad\textbf{+ DensiTok} & \textbf{25.31} & \textbf{0.866} & \textbf{0.155} & \textbf{45.8} & \textbf{84.7} & \textbf{0.884} & \textbf{0.107} & \textbf{25.17} & \textbf{0.851} & \textbf{0.152} & \textbf{67.3} & \textbf{93.3} & \textbf{0.832} & \textbf{0.137} & 0.361 \\
        \bottomrule
      \end{tabular}
    }
    \label{tab:main_results}
  \end{center}
\end{table}

\section{Experiments}

\subsection{Setup}
\label{sec:exp_setup}

\paragraph{Datasets.}
We train and evaluate DensiTok separately on RealEstate10K~\citep{zhou2018stereo} and DL3DV-10K~\citep{ling2024dl3dv}, using ordered sequences of $N=16$ view in the main setting. 
During training, we sample views from both datasets following WorldMirror~\citep{liu2025worldmirror}. 
For evaluation, we use fixed sequences spanning three difficulty levels based on endpoint image overlap, with nested sets of $n\in\{2,3,5,7\}$ condition views. 

\paragraph{Implementation Details.}
We train DensiTok separately for three pretrained backbones: AnySplat~\citep{jiang2025anysplat}, WorldMirror~\citep{liu2025worldmirror}, and Depth Anything 3~\citep{lin2025depth}. 
Following WorldMirror, we randomly sample a patch-aligned input resolution for each training batch. 
We evaluate the models at a fixed resolution of $294\times518$ pixels. 
Also, we randomly sample training sequences, draw $n$ uniformly from $\{2,\ldots,8\}$, and select the $n$ observed views uniformly without replacement. 
We train LAM-VAE and the DiT for 200k iterations each, then finetune the DiT for one-step inference for an additional 20k iterations. 
All training and evaluation are conducted on four NVIDIA RTX 6000 Ada GPUs. 
Other training details are provided in Appendix~\ref{sec:app_training}.

\paragraph{Evaluation Metrics.}
Since DensiTok enhances geometry tokens shared by the Gaussian, camera, and depth heads, we evaluate all three outputs. We use PSNR, SSIM, and LPIPS for rendering quality. 
Following the evaluation protocols of VGGT~\citep{wang2025vggt} and VGGT-$\Omega$~\citep{wang2026vggtomega}, we measure camera pose accuracy using AUC at $3^\circ$ and $30^\circ$ and depth accuracy using $\delta_{1.25}$ and absolute relative error (AbsRel). 
Since RE10K and DL3DV do not provide official ground-truth depth, we use MoGe-2~\citep{wang2025moge2} predictions as pseudo-labels for depth evaluation.

\subsection{Results}
\label{sec:results}

\begin{figure}[t]
  \centering
  \includegraphics[width=\textwidth]{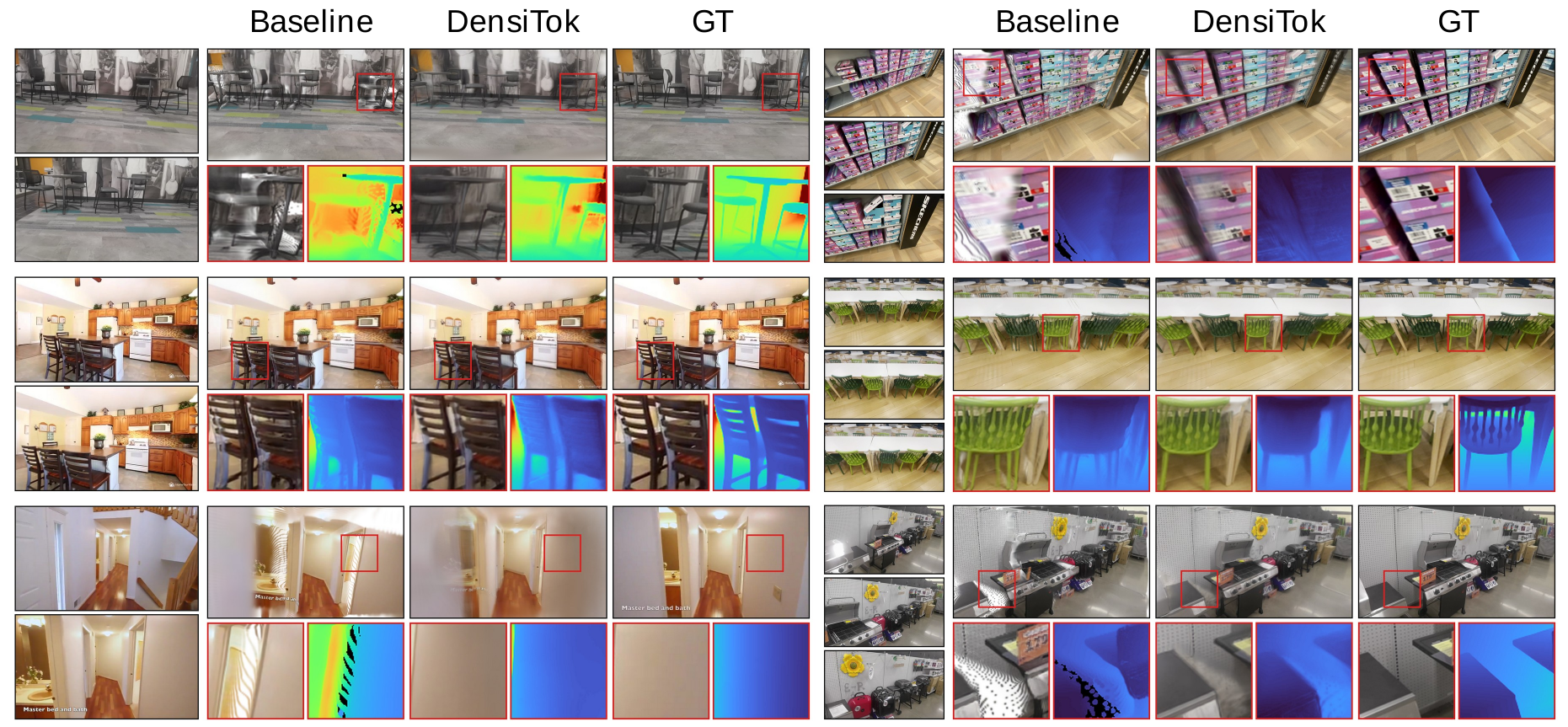}
  \caption{Qualitative novel-view RGB and depth comparisons for AnySplat with and without DensiTok on RE10K and DL3DV, using $n=2$ input views (left) and $n=3$ input views (right).}
  \label{fig:qualitative_results}
\end{figure}

\paragraph{Quantitative results.}
\cref{tab:main_results} compares appearance quality, camera pose accuracy, depth accuracy, and runtime for AnySplat and WorldMirror on RE10K and DL3DV with $n\in\{2,3,5,7\}$ sparse input views and dense-view references with $N=16$. 
Corresponding results for DA3 are available in Appendix \cref{tab:app_more_quantitative}. 
For all appearance evaluations, we follow the scale-alignment procedure used by AnySplat~\citep{jiang2025anysplat}. 
DensiTok improves all three appearance metrics for every backbone, dataset, and evaluated sparse-view setting. 
With only $n=3$ inputs, AnySplat with DensiTok also surpasses the $N=16$ AnySplat reference in all three appearance metrics on both datasets.
DensiTok consistently improves depth estimation across all two backbones, both datasets, and all evaluated sparse-view settings. 
Camera estimation also improves in most cases.
These results suggest that completing the shared geometry representation benefits both appearance and geometry.

\paragraph{Qualitative results.}
\cref{fig:qualitative_results} compares novel-view RGB and depth results under sparse-view conditions on RE10K and DL3DV using AnySplat with and without DensiTok. 
In challenging scenes, DensiTok improves alignment with the target views, geometric reconstruction, and RGB rendering quality. 
This is particularly evident with only $n=2$ inputs, where the baseline leaves missing or fragmented structures. 
DensiTok completes geometric information in these gaps and occluded regions, yielding more coherent surfaces and clearer object boundaries.
Additional qualitative results for other backbones are provided in Appendix~\ref{sec:app_additional_qualitative}.

\subsection{Ablation Studies}
\label{sec:ablations}

\paragraph{Effect of LAM-VAE encoder design.}
\cref{tab:fusion_residual} compares LAM-VAE encoder designs with $n=2$ observed views. 
Uniform fusion fixes the gating weights to $\alpha_\ell=0.25$ for all four encoder levels, while MLP fusion compresses concatenated level features with a simple MLP. 
We evaluate level-adaptive fusion and the level-wise residual separately and together relative to the uniform fusion baseline. 
Each component improves on uniform and MLP fusion, while combining them yields the best results across all metrics on both datasets. 
This supports their complementary roles.

\begin{table}[!t]
  \centering
  \caption{Ablations of Level-Adaptive Modulation with AnySplat and $n=2$ observed views.}
  \vspace{+2mm}
  \label{tab:fusion_residual}
  \begingroup
  \scriptsize
  \setlength{\tabcolsep}{2pt}
  \renewcommand{\arraystretch}{1.15}
  \resizebox{\textwidth}{!}{%
    \begin{tabular}{@{}l*{7}{c}@{\hspace{6pt}}*{7}{c}@{}}
      \toprule
      & \multicolumn{7}{c}{\textbf{RE10K}}
        & \multicolumn{7}{c}{\textbf{DL3DV}} \\
      \cmidrule(lr){2-8}\cmidrule(lr){9-15}
      LAM-VAE encoder
        & PSNR & SSIM & LPIPS & AUC@$3^{\circ}$ & AUC@$30^{\circ}$ & $\delta_{1.25}$ & AbsRel           & PSNR & SSIM & LPIPS & AUC@$3^{\circ}$ & AUC@$30^{\circ}$ & $\delta_{1.25}$ & AbsRel \\
      \midrule
      Uniform fusion & 19.76 & 0.673 & 0.266 & 18.3 & 76.3 & 0.808 & 0.184 & 19.53 & 0.611 & 0.314 & 21.7 & 85.2 & 0.760 & 0.199 \\
      MLP fusion & 21.19 & 0.725 & 0.243 & 20.8 & 77.4 & 0.831 & 0.164 & 20.53 & 0.644 & 0.291 & 23.8 & 86.1 & 0.781 & 0.181 \\
      + level-adaptive fusion & 22.11 & 0.756 & 0.228 & 25.0 & 78.1 & 0.836 & 0.159 & 21.89 & 0.684 & 0.267 & 28.8 & 87.6 & 0.787 & 0.177 \\
      + level-wise residual & 21.86 & 0.747 & 0.234 & 24.2 & 78.6 & 0.841 & 0.156 & 21.56 & 0.675 & 0.274 & 27.5 & 88.2 & 0.794 & 0.173 \\
      + both & \textbf{22.56} & \textbf{0.766} & \textbf{0.223} & \textbf{26.7} & \textbf{78.9} & \textbf{0.846} & \textbf{0.153} & \textbf{22.40} & \textbf{0.699} & \textbf{0.255} & \textbf{30.4} & \textbf{88.9} & \textbf{0.801} & \textbf{0.170} \\
      \bottomrule
    \end{tabular}%
  }
  \endgroup

  \par\vspace{0.8em}
  \caption{Effect of the total number of views $N$ with AnySplat and $n=2$ observed views.}
\vspace{+2mm}
  \label{tab:view_count}
  \begingroup
  \scriptsize
  \setlength{\tabcolsep}{1.5pt}
  \renewcommand{\arraystretch}{1.15}
  \resizebox{0.9\textwidth}{!}{%
    \begin{tabular}{@{}c*{5}{c}@{\hspace{6pt}}*{5}{c}@{\hspace{6pt}}cc@{}}
      \toprule
      & \multicolumn{5}{c}{\textbf{RE10K}}
        & \multicolumn{5}{c}{\textbf{DL3DV}}
        & \multicolumn{2}{c}{\textbf{Efficiency}} \\
      \cmidrule(lr){2-6}\cmidrule(lr){7-11}\cmidrule(lr){12-13}
      Total views~$(N)$
        & PSNR & SSIM & LPIPS & AUC$@3^\circ\!$ & AUC$@30^\circ\!$
        & PSNR & SSIM & LPIPS & AUC$@3^\circ\!$ & AUC$@30^\circ\!$
        & Runtime & Peak GPU mem. \\
      \midrule
      2 & 20.94 & 0.714 & 0.245 & 20.8 & 77.5 & 20.03 & 0.631 & 0.278 & 22.9 & 86.3 & \textbf{0.109} & \textbf{5.527} \\
      4 & 22.08 & 0.757 & 0.239 & 22.9 & 78.1 & 21.28 & 0.636 & 0.257 & 23.3 & 87.6 & 0.116 & 6.196 \\
      8 & 22.49 & 0.762 & 0.231 & 25.4 & 78.6 & 22.18 & 0.687 & \textbf{0.251} & 28.8 & 89.1 & 0.158 & 7.175 \\
      12 & 22.31 & \textbf{0.768} & 0.226 & 25.0 & 78.8 & 22.37 & 0.695 & 0.253 & 30.4 & \textbf{89.3} & 0.209 & 8.407 \\
      16 & \textbf{22.56} & 0.766 & \textbf{0.223} & \textbf{26.7} & \textbf{78.9} & \textbf{22.40} & \textbf{0.699} & 0.255 & \textbf{30.4} & 88.9 & 0.258 & 9.471 \\
      \bottomrule
    \end{tabular}%
  }
  \endgroup
\end{table}

\paragraph{Effect of the number of views.}
To examine how the number of completed views affects reconstruction quality and efficiency, we vary the total number of views $N$ with $n=2$ observed views (\cref{tab:view_count}).
At $N=n=2$, no tokens are generated for unobserved views, allowing us to assess the effect of refining the original geometry tokens through LAM-VAE.
The limited gains over the original AnySplat baseline indicate that most of DensiTok's improvement comes from generating additional geometry tokens.
 Increasing $N$ generates tokens for more unobserved views and generally improves reconstruction quality, at the cost of longer runtime and higher peak GPU memory. 
 The additional views provide the reconstruction heads with geometry tokens for a denser set of viewpoints while keeping the observed inputs fixed. 
 In our experiments, gains diminish beyond $N=12$, making the tradeoff between quality and computational cost less favorable.

\section{Limitations and Future Work}
\label{sec_limitations_future_work}
DensiTok currently restricts geometry token augmentation to view interpolation and does not implement view extrapolation beyond the observed camera interval.
Virtual cameras are interpolated between the first and last observed cameras, and our evaluation likewise considers novel views within this interval.
Our results therefore do not establish reconstruction quality for extrapolated viewpoints.
Future work could extend virtual camera sampling beyond the observed trajectory and adapt latent completion to the weaker visual constraints at these viewpoints.
A key challenge is to infer newly visible geometry and appearance while maintaining consistency with the observed scene.

\section{Conclusion}
\label{sec:conclusion}
DensiTok improves unposed feed-forward 3D Gaussian Splatting by completing the geometry tokens of unobserved views.
LAM-VAE compresses multi-level tokens, and a camera-conditioned DiT completes missing-view latents in one step.
Across three backbones on RealEstate10K and DL3DV, DensiTok improves rendering, camera, and depth accuracy, with sparse inputs surpassing dense-view references in some settings.

\bibliography{iclr2027_conference}

@inproceedings{yao2026flow2gan,
  title={Flow2GAN: Hybrid flow matching and GAN with multi-resolution network for few-step high-fidelity audio generation},
  author={Yao, Zengwei and Kang, Wei and Zhu, Han and Guo, Liyong and Ye, Lingxuan and Kuang, Fangjun and Zhuang, Weiji and Li, Zhaoqing and Han, Zhifeng and Lin, Long and others},
  booktitle={International Conference on Learning Representations},
  volume={2026},
  pages={747--761},
  year={2026}
}

@article{chen2026multi,
  title={Multi-Fidelity Flow Matching: Cascaded Refinement of PDE Solutions},
  author={Chen, Sipeng and Liu, Junliang and Tang, Hewei and Li, Shibo},
  journal={arXiv preprint arXiv:2605.16118},
  year={2026}
}

@inproceedings{xu2025freesplatter,
  title={Freesplatter: Pose-free gaussian splatting for sparse-view 3d reconstruction},
  author={Xu, Jiale and Gao, Shenghua and Shan, Ying},
  booktitle={2025 IEEE/CVF International Conference on Computer Vision (ICCV)},
  pages={25442--25452},
  year={2025},
  organization={IEEE}
}

@inproceedings{hong2025pf3plat,
  title={PF3plat: Pose-free feed-forward 3D Gaussian splatting for novel view synthesis},
  author={Hong, Sunghwan and Jung, Jaewoo and Shin, Heeseong and Han, Jisang and Yang, Jiaolong and Luo, Chong and Kim, Seungryong},
  booktitle={Forty-second International Conference on Machine Learning},
  year={2025}
}

@article{ba2016layer,
  title={Layer normalization},
  author={Ba, Jimmy Lei and Kiros, Jamie Ryan and Hinton, Geoffrey E},
  journal={arXiv preprint arXiv:1607.06450},
  year={2016}
}

@article{kingma2013auto,
  title={Auto-encoding variational bayes},
  author={Kingma, Diederik P and Welling, Max},
  journal={arXiv preprint arXiv:1312.6114},
  year={2013}
}

@article{su2024roformer,
  title={Roformer: Enhanced transformer with rotary position embedding},
  author={Su, Jianlin and Ahmed, Murtadha and Lu, Yu and Pan, Shengfeng and Bo, Wen and Liu, Yunfeng},
  journal={Neurocomputing},
  volume={568},
  pages={127063},
  year={2024},
  publisher={Elsevier}
}

@inproceedings{wang2025vggt,
  title={Vggt: Visual geometry grounded transformer},
  author={Wang, Jianyuan and Chen, Minghao and Karaev, Nikita and Vedaldi, Andrea and Rupprecht, Christian and Novotny, David},
  booktitle={2025 IEEE/CVF Conference on Computer Vision and Pattern Recognition (CVPR)},
  pages={5294--5306},
  year={2025},
  organization={IEEE}
}

@InProceedings{wang2026vggtomega,
    author    = {Wang, Jianyuan and Chen, Minghao and Zhang, Shangzhan and Karaev, Nikita and Sch\"onberger, Johannes and Labatut, Patrick and Bojanowski, Piotr and Novotny, David and Vedaldi, Andrea and Rupprecht, Christian},
    title     = {VGGT-\ensuremath{\Omega}},
    booktitle = {Proceedings of the IEEE/CVF Conference on Computer Vision and Pattern Recognition (CVPR)},
    month     = {June},
    year      = {2026},
    pages     = {21486-21499}
}

@inproceedings{liang2025wonderland,
  title={Wonderland: Navigating 3d scenes from a single image},
  author={Liang, Hanwen and Cao, Junli and Goel, Vidit and Qian, Guocheng and Korolev, Sergei and Terzopoulos, Demetri and Plataniotis, Konstantinos N and Tulyakov, Sergey and Ren, Jian},
  booktitle={2025 IEEE/CVF Conference on Computer Vision and Pattern Recognition (CVPR)},
  pages={798--810},
  year={2025},
  organization={IEEE}
}

@inproceedings{yang2025prometheus,
  title={Prometheus: 3d-aware latent diffusion models for feed-forward text-to-3d scene generation},
  author={Yang, Yuanbo and Shao, Jiahao and Li, Xinyang and Shen, Yujun and Geiger, Andreas and Liao, Yiyi},
  booktitle={2025 IEEE/CVF Conference on Computer Vision and Pattern Recognition (CVPR)},
  pages={2857--2869},
  year={2025},
  organization={IEEE}
}

@article{lan2025ln3diff++,
  title={LN3DIFF++: Scalable latent neural fields diffusion for speedy 3D generation},
  author={Lan, Yushi and Hong, Fangzhou and Zhou, Shangchen and Yang, Shuai and Meng, Xuyi and Chen, Yongwei and Lyu, Zhaoyang and Dai, Bo and Pan, Xingang and Loy, Chen Change},
  journal={IEEE Transactions on Pattern Analysis and Machine Intelligence},
  year={2025},
  publisher={IEEE}
}

@article{lin2025depth,
  title={Depth anything 3: Recovering the visual space from any views},
  author={Lin, Haotong and Chen, Sili and Liew, Junhao and Chen, Donny Y and Li, Zhenyu and Shi, Guang and Feng, Jiashi and Kang, Bingyi},
  journal={arXiv preprint arXiv:2511.10647},
  year={2025}
}

@article{liu2025worldmirror,
  title={Worldmirror: Universal 3d world reconstruction with any-prior prompting},
  author={Liu, Yifan and Min, Zhiyuan and Wang, Zhenwei and Wu, Junta and Wang, Tengfei and Yuan, Yixuan and Luo, Yawei and Guo, Chunchao},
  journal={arXiv preprint arXiv:2510.10726},
  year={2025}
}

@article{jiang2025anysplat,
  title={Anysplat: Feed-forward 3d gaussian splatting from unconstrained views},
  author={Jiang, Lihan and Mao, Yucheng and Xu, Linning and Lu, Tao and Ren, Kerui and Jin, Yichen and Xu, Xudong and Yu, Mulin and Pang, Jiangmiao and Zhao, Feng and others},
  journal={ACM Transactions on Graphics (TOG)},
  volume={44},
  number={6},
  pages={1--16},
  year={2025},
  publisher={ACM New York, NY, USA}
}

@article{lipman2022flow,
  title={Flow matching for generative modeling},
  author={Lipman, Yaron and Chen, Ricky TQ and Ben-Hamu, Heli and Nickel, Maximilian and Le, Matt},
  journal={arXiv preprint arXiv:2210.02747},
  year={2022}
}

@inproceedings{jang2026repurposing,
  title={Repurposing Geometric Foundation Models for Multi-view Diffusion},
  author={Wooseok Jang and Seonghu Jeon and Jisang Han and Jinhyeok Choi and Minkyung Kwon and Seungryong Kim and Saining Xie and Sainan Liu},
  booktitle={European Conference on Computer Vision (ECCV)},
  year={2026}
}

@article{wang2025volsplat,
  title={Volsplat: Rethinking feed-forward 3d gaussian splatting with voxel-aligned prediction},
  author={Wang, Weijie and Chen, Yeqing and Zhang, Zeyu and Liu, Hengyu and Wang, Haoxiao and Feng, Zhiyuan and Qin, Wenkang and Chen, Feng and Bian, Jia-Wang and Zhu, Zheng and others},
  journal={arXiv preprint arXiv:2509.19297},
  year={2025}
}

@inproceedings{huang2025no,
  title={No pose at all: Self-supervised pose-free 3d gaussian splatting from sparse views},
  author={Huang, Ranran and Mikolajczyk, Krystian},
  booktitle={2025 IEEE/CVF International Conference on Computer Vision (ICCV)},
  pages={27947--27957},
  year={2025},
  organization={IEEE}
}

@inproceedings{kang2025selfsplat,
  title={Selfsplat: Pose-free and 3d prior-free generalizable 3d gaussian splatting},
  author={Kang, Gyeongjin and Yoo, Jisang and Park, Jihyeon and Nam, Seungtae and Im, Hyeonsoo and Shin, Sangheon and Kim, Sangpil and Park, Eunbyung},
  booktitle={2025 IEEE/CVF Conference on Computer Vision and Pattern Recognition (CVPR)},
  pages={22012--22022},
  year={2025},
  organization={IEEE}
}

@inproceedings{ye2025no,
  title={No pose, no problem: Surprisingly simple 3d gaussian splats from sparse unposed images},
  author={Ye, Botao and Liu, Sifei and Xu, Haofei and Li, Xueting and Pollefeys, Marc and Yang, Ming-Hsuan and Peng, Songyou},
  booktitle={International Conference on Learning Representations},
  volume={2025},
  pages={54009--54033},
  year={2025}
}

@inproceedings{edstedt2024roma,
  title={Roma: Robust dense feature matching},
  author={Edstedt, Johan and Sun, Qiyu and B{\"o}kman, Georg and Wadenb{\"a}ck, M{\aa}rten and Felsberg, Michael},
  booktitle={2024 IEEE/CVF Conference on Computer Vision and Pattern Recognition (CVPR)},
  pages={19790--19800},
  year={2024},
  organization={IEEE}
}

@inproceedings{xu2025depthsplat,
  title={Depthsplat: Connecting gaussian splatting and depth},
  author={Xu, Haofei and Peng, Songyou and Wang, Fangjinhua and Blum, Hermann and Barath, Daniel and Geiger, Andreas and Pollefeys, Marc},
  booktitle={2025 IEEE/CVF Conference on Computer Vision and Pattern Recognition (CVPR)},
  pages={16453--16463},
  year={2025},
  organization={IEEE}
}

@article{smart2024splatt3r,
  title={Splatt3r: Zero-shot gaussian splatting from uncalibrated image pairs},
  author={Smart, Brandon and Zheng, Chuanxia and Laina, Iro and Prisacariu, Victor Adrian},
  journal={arXiv preprint arXiv:2408.13912},
  year={2024}
}

@inproceedings{ni2026g4splat,
  title={G4splat: Geometry-guided gaussian splatting with generative prior},
  author={Ni, Junfeng and Chen, Yixin and Yang, Zhifei and Liu, Yu and Lu, Ruijie and Zhu, Song-Chun and Huang, Siyuan},
  booktitle={International Conference on Learning Representations},
  volume={2026},
  pages={138924--138951},
  year={2026}
}

@inproceedings{wang2025videoscene,
  title={Videoscene: Distilling video diffusion model to generate 3d scenes in one step},
  author={Wang, Hanyang and Liu, Fangfu and Chi, Jiawei and Duan, Yueqi},
  booktitle={2025 IEEE/CVF Conference on Computer Vision and Pattern Recognition (CVPR)},
  pages={16475--16485},
  year={2025},
  organization={IEEE}
}

@article{yin2025gsfixer,
  title={Gsfixer: Improving 3d gaussian splatting with reference-guided video diffusion priors},
  author={Yin, Xingyilang and Zhang, Qi and Chang, Jiahao and Feng, Ying and Fan, Qingnan and Yang, Xi and Pun, Chi-Man and Zhang, Huaqi and Cun, Xiaodong},
  journal={arXiv preprint arXiv:2508.09667},
  year={2025}
}

@article{chen2024mvsplat360,
  title={Mvsplat360: Feed-forward 360 scene synthesis from sparse views},
  author={Chen, Yuedong and Zheng, Chuanxia and Xu, Haofei and Zhuang, Bohan and Vedaldi, Andrea and Cham, Tat-Jen and Cai, Jianfei},
  journal={Advances in Neural Information Processing Systems},
  volume={37},
  pages={107064--107086},
  year={2024}
}

@inproceedings{fischer2025flowr,
  title={Flowr: Flowing from sparse to dense 3d reconstructions},
  author={Fischer, Tobias and Bul{\`o}, Samuel Rota and Yang, Yung-Hsu and Keetha, Nikhil and Porzi, Lorenzo and M{\"u}ller, Norman and Schwarz, Katja and Luiten, Jonathon and Pollefeys, Marc and Kontschieder, Peter},
  booktitle={2025 IEEE/CVF International Conference on Computer Vision (ICCV)},
  pages={27702--27712},
  year={2025},
  organization={IEEE}
}

@inproceedings{zhong2025taming,
  title={Taming video diffusion prior with scene-grounding guidance for 3d gaussian splatting from sparse inputs},
  author={Zhong, Yingji and Li, Zhihao and Chen, Dave Zhenyu and Hong, Lanqing and Xu, Dan},
  booktitle={2025 IEEE/CVF Conference on Computer Vision and Pattern Recognition (CVPR)},
  pages={6133--6143},
  year={2025},
  organization={IEEE}
}

@inproceedings{szymanowicz2025bolt3d,
  title={Bolt3d: Generating 3d scenes in seconds},
  author={Szymanowicz, Stanislaw and Zhang, Jason Y and Srinivasan, Pratul and Gao, Ruiqi and Brussee, Arthur and Holynski, Aleksander and Martin-Brualla, Ricardo and Barron, Jonathan T and Henzler, Philipp},
  booktitle={2025 IEEE/CVF International Conference on Computer Vision (ICCV)},
  pages={24846--24857},
  year={2025},
  organization={IEEE}
}

@inproceedings{gao2024cat3d,
 author = {Gao, Ruiqi and Ho\l y\'{n}ski, Aleksander and Henzler, Philipp and Brussee, Arthur and Martin-Brualla, Ricardo and Srinivasan, Pratul and Barron, Jonathan T. and Poole, Ben},
 booktitle = {Advances in Neural Information Processing Systems},
 doi = {10.52202/079017-2403},
 editor = {A. Globerson and L. Mackey and D. Belgrave and A. Fan and U. Paquet and J. Tomczak and C. Zhang},
 pages = {75468--75494},
 publisher = {Curran Associates, Inc.},
 title = {CAT3D: Create Anything in 3D with Multi-View Diffusion Models},
 url = {https://proceedings.neurips.cc/paper_files/paper/2024/file/89e4433fec4b99f1d859db57af1e0a0f-Paper-Conference.pdf},
 volume = {37},
 year = {2024}
}

@inproceedings{wang2025vistadream,
  title={Vistadream: Sampling multiview consistent images for single-view scene reconstruction},
  author={Wang, Haiping and Liu, Yuan and Liu, Ziwei and Wang, Wenping and Dong, Zhen and Yang, Bisheng},
  booktitle={2025 IEEE/CVF International Conference on Computer Vision (ICCV)},
  pages={26772--26782},
  year={2025},
  organization={IEEE}
}

@inproceedings{charatan2024pixelsplat,
  title={pixelsplat: 3d gaussian splats from image pairs for scalable generalizable 3d reconstruction},
  author={Charatan, David and Li, Sizhe Lester and Tagliasacchi, Andrea and Sitzmann, Vincent},
  booktitle={2024 IEEE/CVF Conference on Computer Vision and Pattern Recognition (CVPR)},
  pages={19457--19467},
  year={2024},
  organization={IEEE}
}

@inproceedings{chen2024mvsplat,
  title={Mvsplat: Efficient 3d gaussian splatting from sparse multi-view images},
  author={Chen, Yuedong and Xu, Haofei and Zheng, Chuanxia and Zhuang, Bohan and Pollefeys, Marc and Geiger, Andreas and Cham, Tat-Jen and Cai, Jianfei},
  booktitle={European conference on computer vision},
  pages={370--386},
  year={2024},
  organization={Springer}
}

@inproceedings{peebles2023scalable,
  title={Scalable diffusion models with transformers},
  author={Peebles, William and Xie, Saining},
  booktitle={2023 IEEE/CVF International Conference on Computer Vision (ICCV)},
  pages={4172--4182},
  year={2023},
  organization={IEEE}
}

@article{liu2022flow,
  title={Flow straight and fast: Learning to generate and transfer data with rectified flow},
  author={Liu, Xingchao and Gong, Chengyue and Liu, Qiang},
  journal={arXiv preprint arXiv:2209.03003},
  year={2022}
}

@inproceedings{garcia2025fine,
  title={Fine-tuning image-conditional diffusion models is easier than you think},
  author={Garcia, Gonzalo Martin and Zeid, Karim Abou and Schmidt, Christian and De Geus, Daan and Hermans, Alexander and Leibe, Bastian},
  booktitle={Proceedings of the Winter Conference on Applications of Computer Vision},
  pages={753--762},
  year={2025}
}

@inproceedings{he2025lotus,
  title={Lotus: Diffusion-based visual foundation model for high-quality dense prediction},
  author={He, Jing and Li, Haodong and Yin, Wei and Liang, Yixun and Li, Leheng and Zhou, Kaiqiang and Zhang, Hongbo and Liu, Bingbing and Chen, YingCong},
  booktitle={International Conference on Learning Representations},
  volume={2025},
  pages={89454--89467},
  year={2025}
}

@article{zhou2018stereo,
  title={Stereo magnification: Learning view synthesis using multiplane images},
  author={Zhou, Tinghui and Tucker, Richard and Flynn, John and Fyffe, Graham and Snavely, Noah},
  journal={arXiv preprint arXiv:1805.09817},
  year={2018}
}

@inproceedings{ling2024dl3dv,
  title={Dl3dv-10k: A large-scale scene dataset for deep learning-based 3d vision},
  author={Ling, Lu and Sheng, Yichen and Tu, Zhi and Zhao, Wentian and Xin, Cheng and Wan, Kun and Yu, Lantao and Guo, Qianyu and Yu, Zixun and Lu, Yawen and others},
  booktitle={2024 IEEE/CVF Conference on Computer Vision and Pattern Recognition (CVPR)},
  pages={22160--22169},
  year={2024},
  organization={IEEE}
}

@inproceedings{li2025nvcomposer,
  title={Nvcomposer: Boosting generative novel view synthesis with multiple sparse and unposed images},
  author={Li, Lingen and Zhang, Zhaoyang and Li, Yaowei and Xu, Jiale and Hu, Wenbo and Li, Xiaoyu and Cheng, Weihao and Gu, Jinwei and Xue, Tianfan and Shan, Ying},
  booktitle={2025 IEEE/CVF Conference on Computer Vision and Pattern Recognition (CVPR)},
  pages={777--787},
  year={2025},
  organization={IEEE}
}

@inproceedings{lu2025matrix3d,
  title={Matrix3d: Large photogrammetry model all-in-one},
  author={Lu, Yuanxun and Zhang, Jingyang and Fang, Tian and Nahmias, Jean-Daniel and Tsin, Yanghai and Quan, Long and Cao, Xun and Yao, Yao and Li, Shiwei},
  booktitle={2025 IEEE/CVF Conference on Computer Vision and Pattern Recognition (CVPR)},
  pages={11250--11263},
  year={2025},
  organization={IEEE}
}

@inproceedings{wang2025moge2,
 author = {Wang, Ruicheng and Xu, Sicheng and Dong, Yue and Deng, Yu and Xiang, Jianfeng and Lv, Zelong and Sun, Guangzhong and Tong, Xin and Yang, Jiaolong},
 booktitle = {Advances in Neural Information Processing Systems},
 doi = {10.52202/085713-1207},
 editor = {D. Belgrave and C. Zhang and H. Lin and R. Pascanu and P. Koniusz and M. Ghassemi and N. Chen},
 pages = {35928--35959},
 publisher = {Curran Associates, Inc.},
 title = {MoGe-2: Accurate Monocular Geometry with Metric Scale and Sharp Details},
 url = {https://proceedings.neurips.cc/paper_files/paper/2025/file/336572db3e99930814d6b328d4220cb6-Paper-Conference.pdf},
 volume = {38, Main Conference},
 year = {2025}
}

@inproceedings{schops2017multi,
  title={A multi-view stereo benchmark with high-resolution images and multi-camera videos},
  author={Schops, Thomas and Schonberger, Johannes L and Galliani, Silvano and Sattler, Torsten and Schindler, Konrad and Pollefeys, Marc and Geiger, Andreas},
  booktitle={Proceedings of the IEEE conference on computer vision and pattern recognition},
  pages={3260--3269},
  year={2017}
}

@inproceedings{dai2017scannet,
  title={Scannet: Richly-annotated 3d reconstructions of indoor scenes},
  author={Dai, Angela and Chang, Angel X and Savva, Manolis and Halber, Maciej and Funkhouser, Thomas and Nie{\ss}ner, Matthias},
  booktitle={2017 IEEE conference on computer vision and pattern recognition (CVPR)},
  pages={2432--2443},
  year={2017},
  organization={IEEE}
}

@inproceedings{barron2022mip,
  title={Mip-nerf 360: Unbounded anti-aliased neural radiance fields},
  author={Barron, Jonathan T and Mildenhall, Ben and Verbin, Dor and Srinivasan, Pratul P and Hedman, Peter},
  booktitle={2022 IEEE/CVF Conference on Computer Vision and Pattern Recognition (CVPR)},
  pages={5460--5469},
  year={2022},
  organization={IEEE}
}
\bibliographystyle{iclr2027_conference}

\clearpage
\appendix
\section*{Appendix}
\label{sec:appendix}

\section{Architecture Details}
\label{sec:app_arch}

\subsection{Frame and Global DiT Blocks}
\label{sec:app_dit}

\begin{wrapfigure}{R}{0.48\textwidth}
  \centering
  \includegraphics[width=\linewidth]{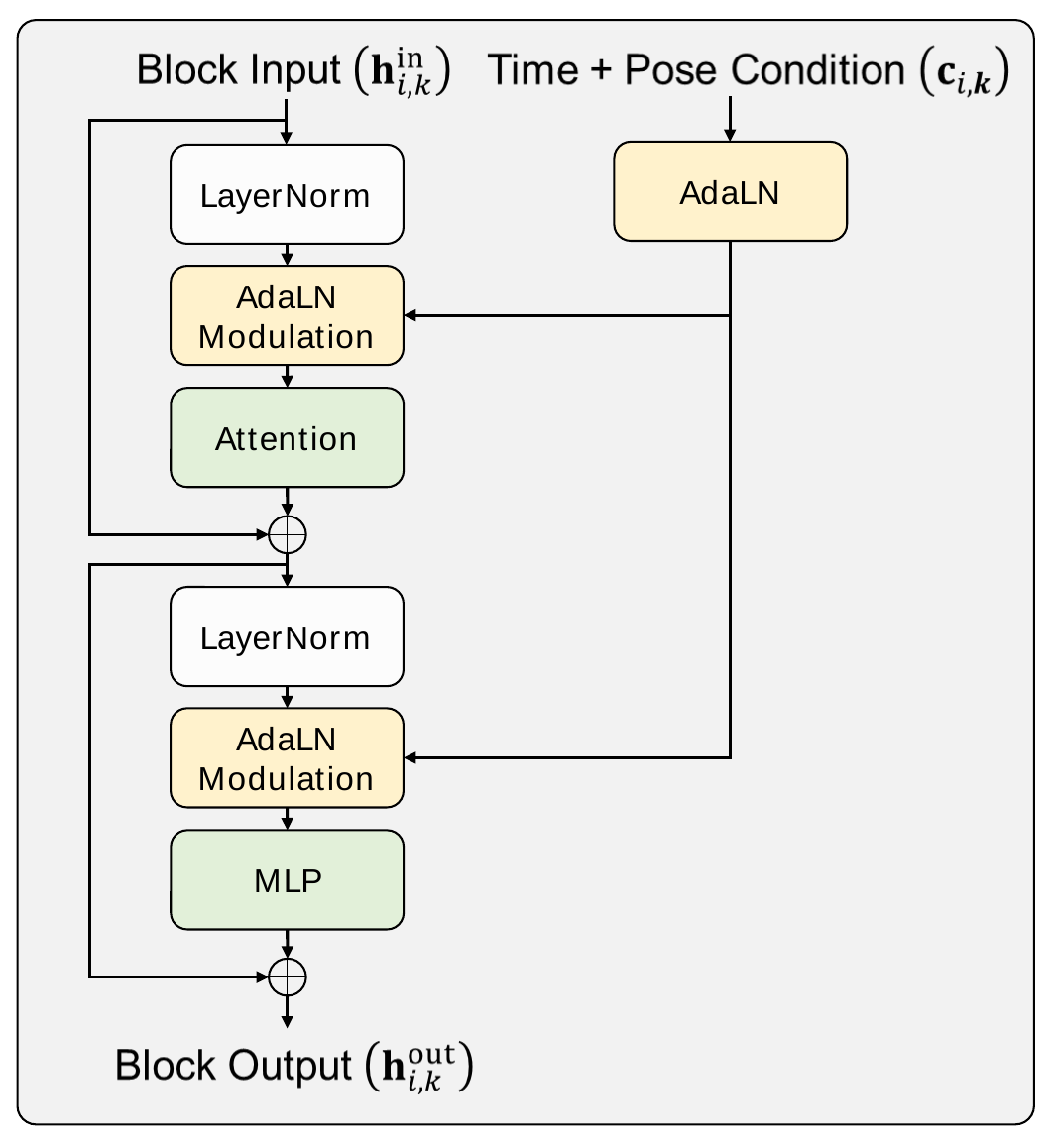}
  \caption{Architecture of a DiT block for geometry latent augmentation. Time and camera embeddings produce per-token AdaLN shift, scale, and gate parameters for the attention and MLP branches. Frame and global blocks differ only in their attention scope.}
  \label{fig:app_dit_block}
  \vspace{-\baselineskip}
\end{wrapfigure}

The DiT in \cref{sec:geometry_latent_augmentation} stacks pairs of blocks, each applying frame attention followed by global attention, as in the VAE encoder transformer (\cref{sec:level_adaptive_modulation}).
Let $d_h$ denote the DiT hidden width.
A frame block attends independently to each view's $K\times d_h$ token matrix, whereas a global block attends jointly to all $N$ views by flattening their tokens into a single $NK\times d_h$ matrix.
Cross-view exchange therefore happens in the global blocks, where noisy unobserved views gather evidence from clean observed views.
Both block types use the architecture in \cref{fig:app_dit_block}, with separate parameters.
For view $i$ and token $k=1,\dots,K$, we denote a block's input and output hidden tokens by $\mathbf{h}^{\mathrm{in}}_{i,k},\mathbf{h}^{\mathrm{out}}_{i,k}\in\mathbb{R}^{d_h}$.

Every block is conditioned through adaptive LayerNorm (AdaLN)~\citep{peebles2023scalable}.
For token $k$ of view $i$, the conditioning vector is $\mathbf{c}_{i,k}=\mathbf{e}^{\mathrm{time}}(\tau_i)+\mathbf{e}^{\mathrm{cam}}_{i,k}\in\mathbb{R}^{d_h}$.
The time embedding $\mathbf{e}^{\mathrm{time}}(\tau_i)$ is shared by all tokens of view $i$, with $\tau_i=t(1-m_i)$: zero for observed views and $t$ for unobserved ones.
The camera embedding $\mathbf{e}^{\mathrm{cam}}_{i,k}$ encodes the Pl\"ucker ray through the patch center, or is a learnable embedding for tokens without a patch location, as described in \cref{sec:geometry_latent_augmentation}.

Each block maps $\mathbf{c}_{i,k}$ to its own shift, scale, and gate vectors for the attention and MLP branches.
This modulation supplies the noise level and camera information, allowing the same blocks to process clean observed views and noisy unobserved views.
After the last global block, a final LayerNorm modulated by the same conditioning vectors and a linear projection back to $d_z$ dimensions produce the predicted flow velocity $\widehat{\mathbf{v}}_{\theta,i}\in\mathbb{R}^{K\times d_z}$ for each view $i$.

\subsection{Appearance Feature Fusion for the Gaussian Head}
\label{sec:app_appfusion}

Some feed-forward backbones, including AnySplat, need an appearance input in addition to the geometry tokens.
Their Gaussian reconstruction head $\mathcal{H}_{\mathrm{G}}$ embeds each view's RGB image with a convolutional image embedding layer $h_{\mathrm{img}}$ and adds the result to its intermediate features, supplying high-frequency appearance for the Gaussian colors.
This assumes an image for every view.
In our setting, however, the $(N-n)$ unobserved views have completed geometry tokens but no RGB images at inference (\cref{sec:geometry_latent_augmentation}).
We therefore supply their missing appearance input by warping the $n$ observed images into these views.

\paragraph{Warped color prior.}
The pretrained camera and depth heads decode $(\tilde{\mathbf{F}}_1, \dots, \tilde{\mathbf{F}}_L)$ into cameras and depth maps for all $N$ views.
We denote these outputs by $\hat{\mathbf{g}}^{\mathrm{dec}}_i$ and $\hat{\mathbf{D}}_i$, distinguishing the decoded cameras from the cameras that condition the DiT.
For an unobserved view $i$ with $m_i=0$, each pixel $\mathbf{u}$ is first unprojected using its predicted camera and depth to a 3D point $\mathbf{y}_i(\mathbf{u})$.
We then reproject this point into an observed view $c$ with $m_c=1$ to obtain the warped color $\mathcal{W}_{c\rightarrow i}(\mathbf{u})=I_c\bigl(\pi_c(\mathbf{y}_i(\mathbf{u}))\bigr)$, where $\pi_c$ is the projection defined by $\hat{\mathbf{g}}^{\mathrm{dec}}_c$.
A binary indicator $\nu_{c\rightarrow i}(\mathbf{u})$ accepts a sample only if the reprojection falls inside view $c$ with positive depth and agrees, up to a relative tolerance, with $\hat{\mathbf{D}}_c$, rejecting occluded samples.
Averaging the accepted colors gives a per-pixel color prior and its validity mask,
\begin{equation}
\begin{aligned}
\mathbf{C}^{\mathrm{prior}}_i(\mathbf{u})
&=\frac{\sum_{c:m_c=1}\nu_{c\rightarrow i}(\mathbf{u})\,\mathcal{W}_{c\rightarrow i}(\mathbf{u})}{\max\bigl(1,\sum_{c:m_c=1}\nu_{c\rightarrow i}(\mathbf{u})\bigr)},\\
M^{\mathrm{prior}}_i(\mathbf{u})
&=\mathds{1}\Bigl[\textstyle\sum_{c:m_c=1}\nu_{c\rightarrow i}(\mathbf{u})>0\Bigr].
\end{aligned}
\label{eq:color-prior}
\end{equation}
This uses only the $n$ observed images at inference.
The cameras and depths of all $N$ views come from the completed tokens, which are themselves produced from the observed views.
No unobserved RGB image is therefore needed.

\paragraph{Fusion and finetuning.}
Let $J_i$ denote the RGB appearance input to the image embedding layer $h_{\mathrm{img}}$ for view $i$.
The appearance validity mask $M^{\mathrm{app}}_i(\mathbf{u})\in\{0,1\}$ equals one if $J_i$ provides a valid RGB value at pixel $\mathbf{u}$ and zero otherwise.
For an observed view ($m_i=1$), the appearance input is $J_i=I_i$ and every pixel is valid, $M^{\mathrm{app}}_i(\mathbf{u})=1$.
For an unobserved view ($m_i=0$), we use $J_i=\mathbf{C}^{\mathrm{prior}}_i$ and $M^{\mathrm{app}}_i(\mathbf{u})=M^{\mathrm{prior}}_i(\mathbf{u})$.
Pixels with no valid warped color use a learned missing-view embedding $\mathbf{e}^{\mathrm{miss}}$ in place of the image embedding.
The appearance feature map is thus
\begin{equation}
\mathbf{F}^{\mathrm{app}}_i(\mathbf{u})
=M^{\mathrm{app}}_i(\mathbf{u})\,h_{\mathrm{img}}(J_i)(\mathbf{u})
+\bigl(1-M^{\mathrm{app}}_i(\mathbf{u})\bigr)\,\mathbf{e}^{\mathrm{miss}},
\label{eq:app-fusion}
\end{equation}
where the same vector $\mathbf{e}^{\mathrm{miss}}$ is used at every invalid pixel.
The map $\mathbf{F}^{\mathrm{app}}_i$ enters $\mathcal{H}_{\mathrm{G}}$ where the embedded RGB image would normally be added.
These holes are out of distribution for the pretrained head, which was trained on complete images, and can corrupt colors if the appearance input is left unadapted.
We therefore finetune $h_{\mathrm{img}}$ and learn $\mathbf{e}^{\mathrm{miss}}$ under the Stage 2 rendering losses (Appendix~\ref{sec:app_training}), so that the Gaussian colors at holes can be inferred from the surrounding observed appearance rather than from the placeholder alone.

\subsection{Model Configurations}
\label{sec:app_model_configurations}

Table~\ref{tab:architecture_config} summarizes the VAE and DiT configurations shared across all three backbones on both datasets. The latent width is $d_z=4$ by default and $d_z=16$ in the latent channel ablation, with unchanged hidden widths and block counts.

Within LAM-VAE, level-adaptive fusion uses a shared projection to $64$ channels and a gating MLP with hidden width $128$. We extract four feature levels at zero-based encoder indices $4,11,17,23$ for AnySplat and WorldMirror, and $19,27,33,39$ for DA3. Each selected feature level has $2048$ channels for AnySplat and WorldMirror, and $3072$ channels for DA3.

\begin{table}[t]
  \centering
  \caption{Shared VAE and DiT architecture for all three backbones. Each pair contains a frame block followed by a global block.}
  \label{tab:architecture_config}
  \begingroup
  \small
  \setlength{\tabcolsep}{8pt}
  \renewcommand{\arraystretch}{1.15}
  \begin{tabular}{@{}lccc@{}}
    \toprule
    \textbf{Setting} & \textbf{VAE encoder} & \textbf{VAE decoder} & \textbf{DiT} \\
    \midrule
    Hidden channels & 512 & 512 & 768 \\
    Frame/global block pairs & 6 & 6 & 12 \\
    Total transformer blocks & 12 & 12 & 24 \\
    Attention heads & 8 & 8 & 12 \\
    \bottomrule
  \end{tabular}
  \endgroup
\end{table}

\section{Training and Evaluation Details}
\label{sec:app_training}

\paragraph{Stage 1. VAE training.}
As described in \cref{sec:overall_training_strategy}, we train the VAE with Level-Adaptive Modulation while keeping the geometry encoder and the pretrained reconstruction heads frozen.
Given the geometry tokens $(\mathbf{F}_1,\dots,\mathbf{F}_L)$ of an input sequence, the VAE samples a latent and decodes it back into tokens at all $L$ levels.
We denote the reconstructed tokens at level $\ell$ by $\hat{\mathbf{F}}_\ell$.
The primary objective is a level-wise normalized reconstruction error,
\begin{equation}
\mathcal{L}_{\mathrm{feat}}=\frac1L\sum_{\ell=1}^{L}\Bigl\|\operatorname{LN}_\ell\bigl(\hat{\mathbf{F}}_\ell\bigr)-\operatorname{LN}_\ell\bigl(\mathbf{F}_\ell\bigr)\Bigr\|_2^2,
\label{eq:loss-feat}
\end{equation}
where $\operatorname{LN}_\ell$ denotes LayerNorm at level $\ell$.
This normalization balances token magnitudes across levels.
We also use the KL divergence $\mathcal{L}_{\mathrm{KL}}$ between the posterior and a standard Gaussian prior.

Token reconstruction alone, however, does not ensure that the pretrained heads can use the reconstructed tokens.
We therefore pass $(\hat{\mathbf{F}}_1,\dots,\hat{\mathbf{F}}_L)$ through the frozen heads and compare their outputs with those obtained from the original tokens $(\mathbf{F}_1,\dots,\mathbf{F}_L)$ without the VAE.
The losses $\mathcal{L}_{\mathrm{rgb}}$ and $\mathcal{L}_{\mathrm{LPIPS}}$ measure the mean squared error and LPIPS distance between the two sets of rendered images.
The losses $\mathcal{L}_{\mathrm{pose}}$ and $\mathcal{L}_{\mathrm{depth}}$ match the predicted cameras and depths.
Together, these losses train the VAE to preserve the reconstruction produced by the backbone.
The total objective is
\begin{equation}
\mathcal{L}_{\mathrm{VAE}}=\lambda_{\mathrm{feat}}\mathcal{L}_{\mathrm{feat}}+\lambda_{\mathrm{KL}}\mathcal{L}_{\mathrm{KL}}+\lambda_{\mathrm{rgb}}\mathcal{L}_{\mathrm{rgb}}+\lambda_{\mathrm{perc}}\mathcal{L}_{\mathrm{LPIPS}}+\lambda_{\mathrm{pose}}\mathcal{L}_{\mathrm{pose}}+\lambda_{\mathrm{depth}}\mathcal{L}_{\mathrm{depth}},
\label{eq:loss-lam}
\end{equation}
with $(\lambda_{\mathrm{feat}},\lambda_{\mathrm{KL}},\lambda_{\mathrm{rgb}},\lambda_{\mathrm{perc}},\lambda_{\mathrm{pose}},\lambda_{\mathrm{depth}})=(0.1,10^{-3},1,0.05,10,1)$ in our main setting.
The KL weight increases linearly from $0$ to $10^{-3}$ over the first $10{,}000$ training steps.
The KL loss uses a free-bits threshold of $0.5$ nats per latent dimension.

\paragraph{Stage 2. DiT training.}
We train the DiT for geometry latent augmentation and its camera-conditioning embeddings.
We also finetune the Gaussian head's image embedding layer $h_{\mathrm{img}}$ and learn the missing-view embedding $\mathbf{e}^{\mathrm{miss}}$, as described in \cref{sec:app_appfusion}.
As described in \cref{sec:overall_training_strategy}, the geometry encoder, VAE, and remaining pretrained head weights stay frozen.
For each training scene, we draw a global time $t=\operatorname{sigmoid}(u)$ with $u\sim\mathcal{N}(0,1)$.
As in \cref{eq:flow_state}, observed views retain their clean latents, while unobserved views interpolate between full-view latents and noise,
\begin{equation}
\mathbf{x}_{t,i}=m_i\,\mathbf{z}^{\mathrm{obs}}_i+(1-m_i)\bigl[(1-t)\,\mathbf{z}_i+t\,\boldsymbol{\epsilon}_i\bigr],\qquad
\boldsymbol{\epsilon}_i\sim\mathcal{N}(\mathbf{0},\mathbf{I}),
\label{eq:flow-interp}
\end{equation}
where $\mathbf{z}_i$ and $\mathbf{z}^{\mathrm{obs}}_i$ are the per-view latents in $\mathbf{Z}$ and $\mathbf{Z}^{\mathrm{obs}}$, respectively.
At $t=1$, unobserved views contain pure noise and observed views remain clean, matching the starting state at inference.
The primary objective is the flow-matching loss over unobserved views,
\begin{equation}
\mathcal{L}_{\mathrm{FM}}=\frac{1}{(N-n)\,K\,d_z}\sum_{i:m_i=0}\bigl\|\widehat{\mathbf{v}}_{\theta,i}-\mathbf{v}^{\star}_i\bigr\|_2^2,\qquad
\mathbf{v}^{\star}_i=\boldsymbol{\epsilon}_i-\mathbf{z}_i,
\label{eq:loss-fm}
\end{equation}
where $\widehat{\mathbf{v}}_{\theta,i}\in\mathbb{R}^{K\times d_z}$ is the DiT's predicted velocity for view $i$ and $\mathbf{v}^{\star}_i$ is its training target.
The denominator averages the error over all latent entries of the $(N-n)$ unobserved views.
Observed views are excluded from this loss and stay at time $\tau_i=0$.

The velocity loss, however, measures error only in latent space.
To supervise the reconstruction itself, we estimate the clean latents of each view as $\hat{\mathbf{z}}^{(t)}_{0,i}=\mathbf{x}_{t,i}-\tau_i\,\widehat{\mathbf{v}}_{\theta,i}$, with $\tau_i=t(1-m_i)$.
Here, $\hat{\mathbf{z}}^{(t)}_{0,i}$ denotes the clean latents of view $i$ estimated from the flow state at time $t$.
At every training step, the VAE decoder maps these estimates to completed geometry tokens $(\tilde{\mathbf{F}}_1,\dots,\tilde{\mathbf{F}}_L)$, which are then passed through the reconstruction heads.
We randomly select a set of view indices $\mathcal{U}\subseteq\{1,\dots,N\}$ for supervision.
The losses $\mathcal{L}_{\mathrm{rgb}}$ and $\mathcal{L}_{\mathrm{LPIPS}}$ compare images rendered from the resulting Gaussians at these views with their ground-truth images.

The frozen camera and depth heads applied to the original full-view tokens $(\mathbf{F}_1,\dots,\mathbf{F}_L)$ also provide teacher predictions for the same views.
The pose loss $\mathcal{L}_{\mathrm{pose}}$ is a Huber loss between the pose encodings predicted from the completed tokens and the teacher's pose encodings at every refinement iteration of the camera head.
Earlier iterations receive geometrically smaller weights.
The depth loss $\mathcal{L}_{\mathrm{depth}}$ is the squared depth error on pixels selected by the teacher's depth confidence.
The total objective is
\begin{equation}
\mathcal{L}_{\mathrm{DiT}}=\lambda_{\mathrm{FM}}\mathcal{L}_{\mathrm{FM}}+\lambda_{\mathrm{rgb}}\mathcal{L}_{\mathrm{rgb}}+\lambda_{\mathrm{perc}}\mathcal{L}_{\mathrm{LPIPS}}+\lambda_{\mathrm{pose}}\mathcal{L}_{\mathrm{pose}}+\lambda_{\mathrm{depth}}\mathcal{L}_{\mathrm{depth}},
\label{eq:loss-glad}
\end{equation}
with $(\lambda_{\mathrm{FM}},\lambda_{\mathrm{rgb}},\lambda_{\mathrm{perc}},\lambda_{\mathrm{pose}},\lambda_{\mathrm{depth}})=(1,1,0.05,10,1)$ in our main setting.

\begin{table}[t]
  \centering
  \caption{Training configurations for the three stages of the main AnySplat experiments.}
  \vspace{+2mm}
  \label{tab:training_config}
  \begingroup
  \footnotesize
  \setlength{\tabcolsep}{5pt}
  \renewcommand{\arraystretch}{1.12}
  \resizebox{0.8\textwidth}{!}{%
    \begin{tabular}{@{}lccc@{}}
      \toprule
      \textbf{Setting} & \textbf{VAE} & \textbf{DiT} & \textbf{One-step finetuning} \\
      \midrule
      Optimizer & AdamW & AdamW & AdamW \\
      Maximum learning rate & $2\times10^{-4}$ & $2\times10^{-4}$ & $4\times10^{-5}$ \\
      Minimum learning rate & $10^{-6}$ & $2\times10^{-6}$ & $2\times10^{-6}$ \\
      Learning-rate schedule & Cosine decay & Cosine decay & Cosine decay \\
      Optimizer betas & $(0.9,0.95)$ & $(0.9,0.95)$ & $(0.9,0.95)$ \\
      Weight decay & $0.01$ & $0.01$ & $0.01$ \\
      Training iterations & $200{,}000$ & $200{,}000$ & $20{,}000$ \\
      Batch size per GPU (sequences) & $\lfloor16/N_{\mathrm{vae}}\rfloor$ & $1$ & $1$ \\
      Mixed precision & BF16 & BF16 & BF16 \\
      LPIPS start iteration & $0$ & $2{,}000$ & $0$ \\
      \bottomrule
    \end{tabular}%
  }
  \endgroup
\end{table}

\paragraph{Stage 3. One-step finetuning.}
Recently, one-step adaptation has been explored in conditional diffusion models~\citep{garcia2025fine, he2025lotus, chen2026multi, yao2026flow2gan}.
Following this principle, we start from the Stage 2 checkpoint and finetune the model for one-step reconstruction by fixing the time of the unobserved views at $t=1$.
We retain the same loss terms and weights as in \cref{eq:loss-glad}, evaluated at this fixed time.
For an unobserved view, a single explicit Euler step from $t=1$ to $t=0$ gives the completed latents $\tilde{\mathbf{z}}_i=\boldsymbol{\epsilon}_i-\widehat{\mathbf{v}}_{\theta,i}$, where $\tilde{\mathbf{z}}_i$ denotes view $i$ in the completed tensor $\tilde{\mathbf{Z}}$.
At this fixed time, the flow-matching loss is therefore equivalent to direct regression onto the full-view latents,
\begin{equation}
\left.\mathcal{L}_{\mathrm{FM}}\right|_{t=1}
=\frac{1}{(N-n)\,K\,d_z}\sum_{i:m_i=0}\bigl\|\tilde{\mathbf{z}}_i-\mathbf{z}_i\bigr\|_2^2.
\label{eq:loss-one-step}
\end{equation}
The rendering losses also supervise the one-step reconstruction end to end.
Together, these losses optimize one-step reconstruction quality without imposing consistency with the Stage 2 model's multi-step ODE solution.

\paragraph{Training configuration.}
Table~\ref{tab:training_config} summarizes the configuration shared by the main AnySplat experiments on RE10K and DL3DV.
Learning rates are effective values for four GPUs, including the square-root scaling with global batch size used in Stages 2 and 3.
Let $N_{\mathrm{vae}}$ denote the number of input views in a Stage 1 training sequence, which can vary during VAE training.
We use $\lfloor16/N_{\mathrm{vae}}\rfloor$ sequences per GPU to keep at most 16 images per GPU.
Stages 2 and 3 use one sequence per GPU, giving a global batch size of 4.
LPIPS is enabled after the first $2{,}000$ iterations in Stage 2 and from the start in Stages 1 and 3.

\paragraph{Evaluation view sampling.}
\label{sec:app_view_sampling}
Following NoPoSplat~\citep{ye2025no}, we estimate image overlap using bidirectional RoMa~\citep{edstedt2024roma} matching.
The overlap between the first and last views determines the difficulty of an interval.
We group intervals into three overlap ranges $(0.55,0.80]$, $(0.30,0.55]$, and $[0.05,0.30]$, with lower overlap indicating less shared visual evidence.
We assign these ranges to sequences with probabilities $25\%$, $50\%$, and $25\%$, respectively, and exclude sequences without a suitable interval in their assigned range.
The resulting evaluation subsets follow these target proportions subject to the availability of eligible sequences.

Within each selected interval, we place $N=16$ reconstruction views along the camera trajectory according to both translation and rotation, and reserve four separate novel views inside the interval for evaluation.
We evaluate 80 sequences for each of RE10K and DL3DV, yielding 320 evaluation images per dataset.
The $n\in\{2,3,5,7\}$ observed views form nested subsets of the $N$ reconstruction views and always include both endpoints.
Increasing $n$ therefore adds observations while retaining the existing ones.
Because estimated image overlap varies with resolution, we sample views separately at each evaluation resolution.

\begin{table}[t]
  \begin{center}
    \caption{Quantitative results for DA3 on RealEstate10K and DL3DV with $n\in\{2,3,5,7\}$ sparse input views and dense-view references ($N=16$). Camera AUC is reported in percent. Runtime is reported in seconds in a shared column.}
    \resizebox{\columnwidth}{!}{%
      \centering
      \setlength{\tabcolsep}{2pt}
      \renewcommand{\arraystretch}{1.2}
      \begin{tabular}{l*{15}{c}}
        \toprule
        \multirow{3}{*}{Methods}
          & \multicolumn{7}{c}{RealEstate10K~\citep{zhou2018stereo}}
          & \multicolumn{7}{c}{DL3DV~\citep{ling2024dl3dv}} & \multirow{2}{*}{Runtime} \\
        \cmidrule(lr){2-8}\cmidrule(lr){9-15}
          & \multicolumn{3}{c}{Appearance} & \multicolumn{2}{c}{Camera} & \multicolumn{2}{c}{Depth}
          & \multicolumn{3}{c}{Appearance} & \multicolumn{2}{c}{Camera} & \multicolumn{2}{c}{Depth} & \\
        \cmidrule(lr){2-4}\cmidrule(lr){5-6}\cmidrule(lr){7-8}
        \cmidrule(lr){9-11}\cmidrule(lr){12-13}\cmidrule(lr){14-15}
          & PSNR$\uparrow$ & SSIM$\uparrow$ & LPIPS$\downarrow$ & AUC@$3^{\circ}\uparrow$ & AUC@$30^{\circ}\uparrow$ & $\delta_{1.25}\uparrow$ & AbsRel$\downarrow$
          & PSNR$\uparrow$ & SSIM$\uparrow$ & LPIPS$\downarrow$ & AUC@$3^{\circ}\uparrow$ & AUC@$30^{\circ}\uparrow$ & $\delta_{1.25}\uparrow$ & AbsRel$\downarrow$ & (s)$\downarrow$ \\
        \midrule
        \multicolumn{16}{l}{\emph{Dense-view} ($N=16$)} \\ [2pt]
        \quad DA3~\citep{lin2025depth} & 21.28 & 0.712 & 0.249 & 43.8 & 86.2 & 0.538 & 0.346 & 22.73 & 0.717 & 0.222 & 87.8 & 97.2 & 0.556 & 0.322 & 0.514 \\
        \midrule
        \multicolumn{16}{l}{\emph{Sparse-view} ($n=2$)} \\ [2pt]
        \quad DA3 & 19.69 & 0.667 & 0.300 & 55.8 & 89.6 & 0.369 & 0.471 & 20.86 & 0.646 & 0.287 & 88.3 & \textbf{98.8} & 0.402 & 0.452 & 0.112 \\
        \quad\textbf{+ DensiTok} & \textbf{22.95} & \textbf{0.791} & \textbf{0.213} & \textbf{61.7} & \textbf{93.2} & \textbf{0.523} & \textbf{0.409} & \textbf{22.16} & \textbf{0.707} & \textbf{0.256} & \textbf{88.8} & \textbf{98.8} & \textbf{0.525} & \textbf{0.387} & 0.255 \\
        \midrule
        \multicolumn{16}{l}{\emph{Sparse-view} ($n=3$)} \\ [2pt]
        \quad DA3 & 20.50 & 0.696 & 0.268 & 52.2 & 89.0 & 0.453 & 0.405 & 21.65 & 0.674 & 0.256 & 88.8 & 97.9 & 0.458 & 0.392 & 0.135 \\
        \quad\textbf{+ DensiTok} & \textbf{25.01} & \textbf{0.841} & \textbf{0.168} & \textbf{64.9} & \textbf{90.4} & \textbf{0.638} & \textbf{0.342} & \textbf{24.23} & \textbf{0.781} & \textbf{0.196} & \textbf{88.9} & \textbf{98.0} & \textbf{0.607} & \textbf{0.330} & 0.308 \\
        \midrule
        \multicolumn{16}{l}{\emph{Sparse-view} ($n=5$)} \\ [2pt]
        \quad DA3 & 20.81 & 0.709 & 0.255 & 49.5 & 87.6 & 0.480 & 0.387 & 22.12 & 0.690 & 0.242 & 86.4 & \textbf{96.9} & 0.485 & 0.363 & 0.187 \\
        \quad\textbf{+ DensiTok} & \textbf{26.58} & \textbf{0.876} & \textbf{0.140} & \textbf{61.6} & \textbf{90.1} & \textbf{0.739} & \textbf{0.256} & \textbf{25.62} & \textbf{0.824} & \textbf{0.163} & \textbf{86.6} & 96.7 & \textbf{0.684} & \textbf{0.278} & 0.333 \\
        \midrule
        \multicolumn{16}{l}{\emph{Sparse-view} ($n=7$)} \\ [2pt]
        \quad DA3 & 21.08 & 0.714 & 0.248 & 47.9 & 87.4 & 0.500 & 0.382 & 22.40 & 0.703 & 0.233 & 85.0 & \textbf{96.2} & 0.496 & 0.350 & 0.244 \\
        \quad\textbf{+ DensiTok} & \textbf{27.19} & \textbf{0.889} & \textbf{0.127} & \textbf{51.1} & \textbf{88.5} & \textbf{0.770} & \textbf{0.212} & \textbf{26.26} & \textbf{0.847} & \textbf{0.147} & \textbf{85.3} & \textbf{96.2} & \textbf{0.717} & \textbf{0.250} & 0.367 \\
        \bottomrule
      \end{tabular}
    }
    \label{tab:app_more_quantitative}
  \end{center}
\end{table}

\section{Additional Results}
\label{sec:app_additional_results}

\subsection{Results with DA3 Baseline}
\label{sec:app_da3}

Table~\ref{tab:app_more_quantitative} evaluates DensiTok with Depth Anything 3 (DA3)~\citep{lin2025depth} on RE10K and DL3DV using the same evaluation protocol as Table~\ref{tab:main_results}. DensiTok consistently improves PSNR, SSIM, and LPIPS across both datasets and all evaluated sparse-view settings. The gains in rendering quality grow as the number of observed views increases.

\begin{figure}[t]
  \centering
  \includegraphics[width=\textwidth]{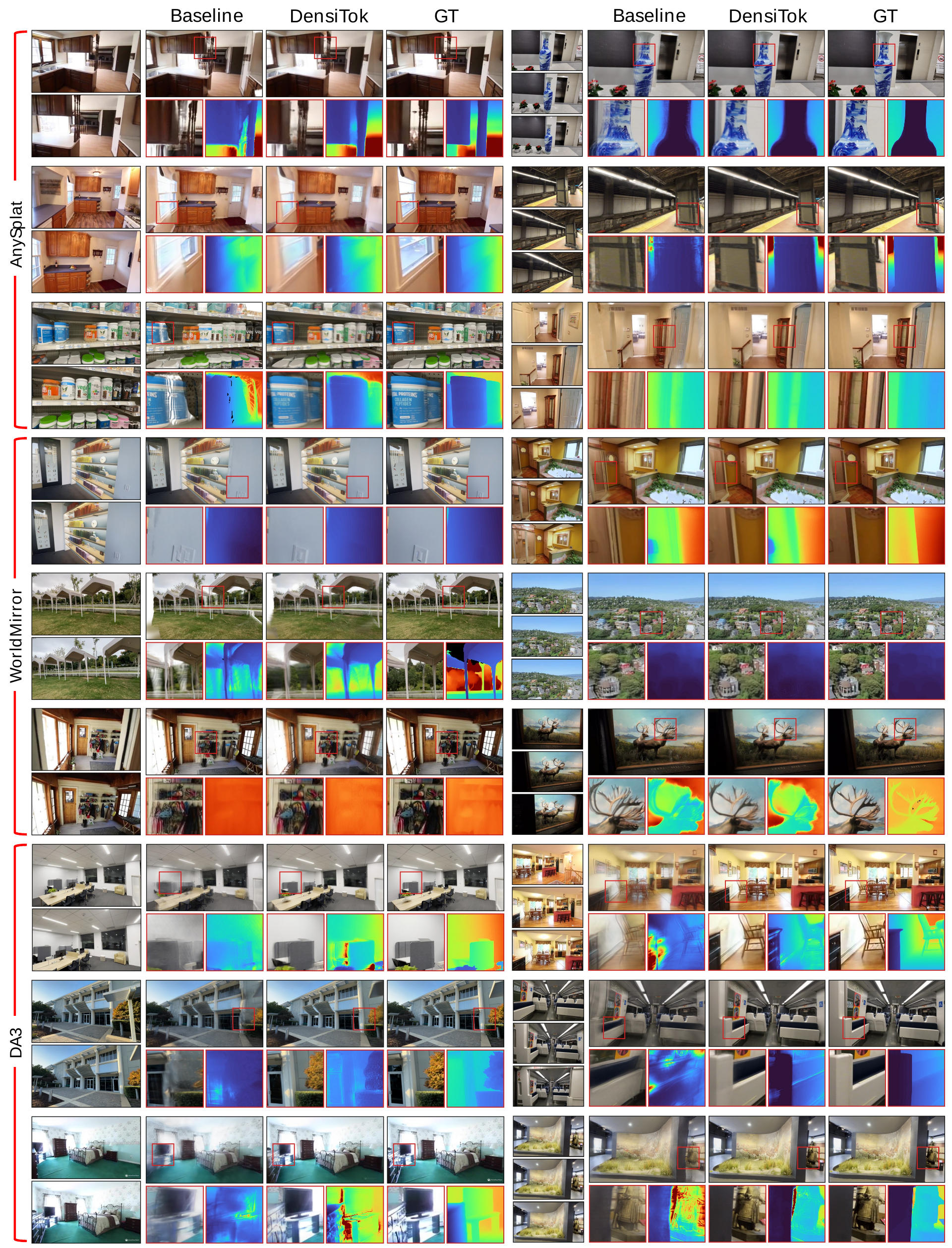}
  \caption{Qualitative novel-view RGB and depth comparisons on RE10K and DL3DV with $n=2$ input views (left) and $n=3$ input views (right), each with and without DensiTok. Red boxes mark the enlarged RGB and depth crops. GT denotes the ground-truth target image.}
  \label{fig:qualitative_results_n2n3}
\end{figure}

\begin{figure}[t]
  \centering
  \includegraphics[width=\textwidth]{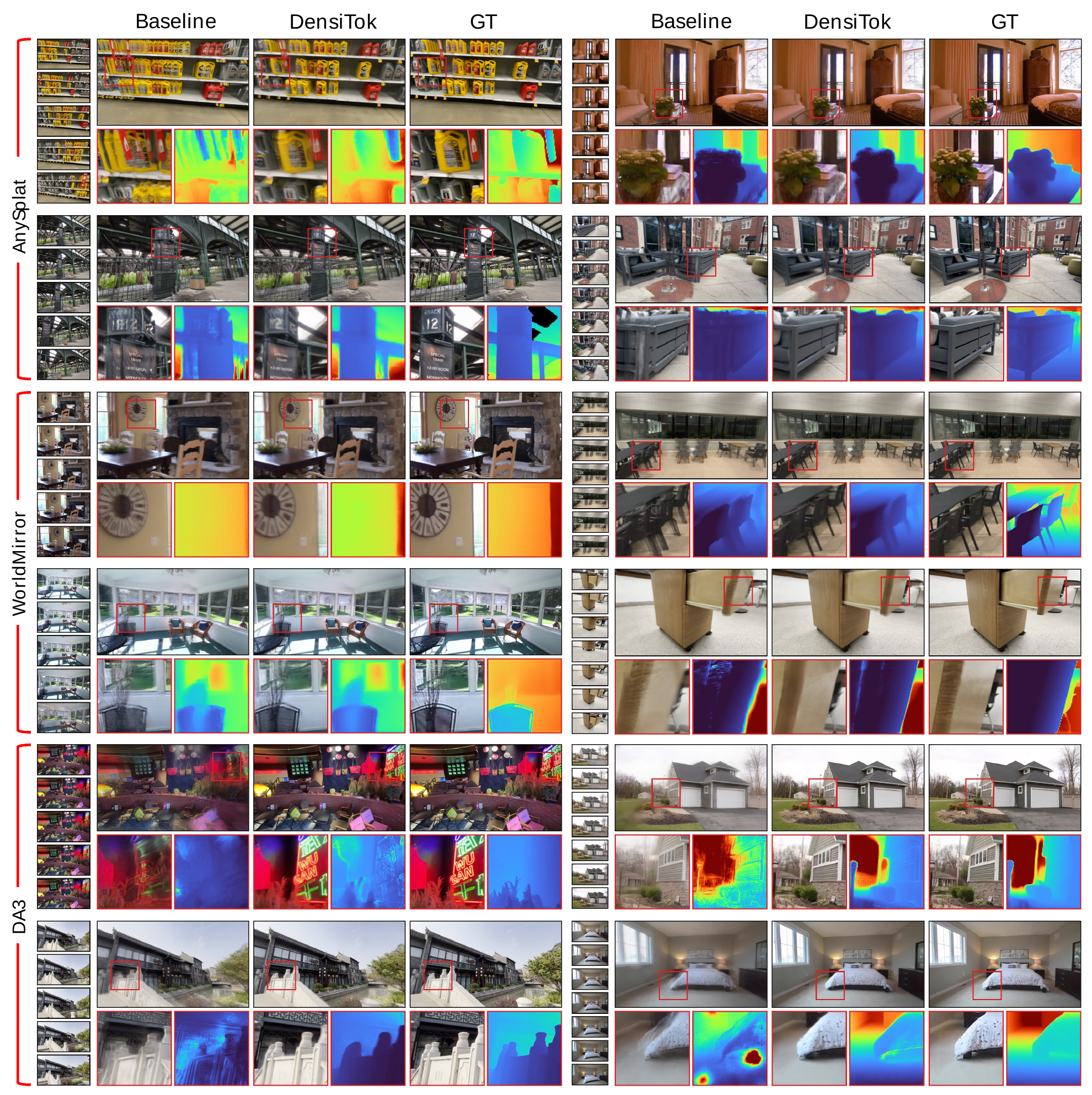}
  \caption{Qualitative novel-view RGB and depth comparisons on RE10K and DL3DV with $n=5$ input views (left) and $n=7$ input views (right), each with and without DensiTok. Red boxes mark the enlarged RGB and depth crops. GT denotes the ground-truth target image.}
  \label{fig:qualitative_results_n5n7}
\end{figure}

Depth estimation also improves consistently on both datasets at every evaluated input count. Camera AUC improves at both angular thresholds on RE10K and at the stricter threshold on DL3DV. At the larger threshold on DL3DV, results remain close to the baseline, with some settings showing no change or a slight decrease.

Even with sparse inputs, DensiTok can surpass the dense-view DA3 reference in all appearance metrics on both datasets. At the larger evaluated input counts, it also exceeds the reference in both depth metrics on both datasets. In these settings, camera AUC is higher than the reference on RE10K but remains lower on DL3DV. DensiTok increases runtime relative to sparse-view DA3, while remaining faster than the dense-view reference across all evaluated settings.

\subsection{Additional Qualitative Results}
\label{sec:app_additional_qualitative}

Figures~\ref{fig:qualitative_results_n2n3} and~\ref{fig:qualitative_results_n5n7} present additional novel-view RGB and depth comparisons on RE10K and DL3DV for AnySplat, WorldMirror, and DA3, each with and without DensiTok. Figure~\ref{fig:qualitative_results_n2n3} shows results with $n=2$ and $n=3$ input views, while Figure~\ref{fig:qualitative_results_n5n7} covers $n=5$ and $n=7$. These examples extend the qualitative comparison in the main paper to all three backbones and all evaluated input counts.

With only two or three observed views, DensiTok reduces RGB blur and produces more coherent depth regions in the highlighted examples, including clearer object contours around monitors and train seats. The benefits remain visible with five or seven views, where the crops show sharper image details and reduced local noise in the depth predictions. These changes are particularly visible in the architectural and furniture examples with DA3.

\subsection{Visualization of Gaussian Reconstructions and Camera Poses}
\label{sec:app_gaussian_camera_visualization}

\begin{figure}[t]
  \centering
  \includegraphics[width=\textwidth]{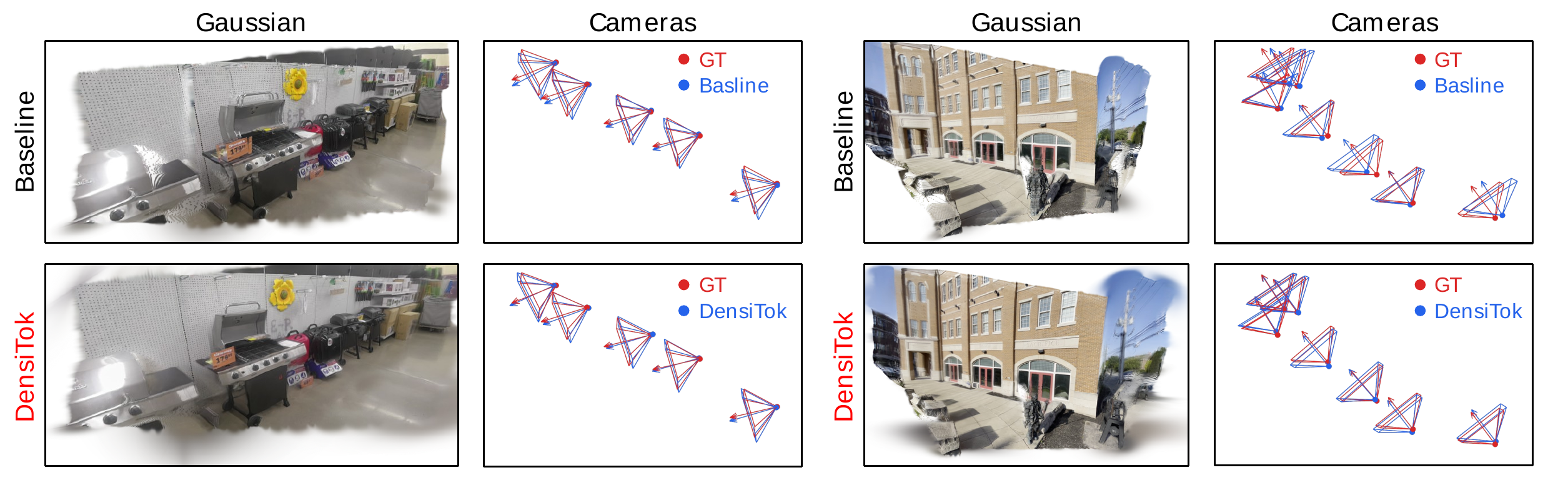}
  \caption{Visualization of Gaussian reconstructions and camera poses for AnySplat with and without DensiTok. The top row shows the original AnySplat backbone and the bottom row shows DensiTok. Each scene is paired with a camera visualization, where red denotes ground truth and blue denotes predictions. DensiTok produces clearer scene structure and closer agreement with ground-truth camera positions and viewing orientations.}
  \label{fig:app_gaussian_camera_visualization}
\end{figure}

\begin{table}[t]
  \centering
  \caption{Comparison with sparse-view feed-forward Gaussian Splatting methods on RE10K and DL3DV at $256\times256$ with $n\in\{2,3,5,7\}$. NoPoSplat results are included for $n\in\{2,3\}$. Intr. and Extr. indicate whether the evaluated configuration requires externally supplied input camera intrinsics and extrinsics for reconstruction. \protect\colorbox{tablebest}{Orange} and \protect\colorbox{tablesecond}{yellow} mark the best and second-best values for each metric at the same $n$, respectively. Rankings are determined before rounding, with ties sharing the same color.}
  \label{tab:app_sparse_comparison}
  \begingroup
  \scriptsize
  \setlength{\tabcolsep}{2pt}
  \renewcommand{\arraystretch}{1.15}
  \resizebox{0.95\textwidth}{!}{%
    \begin{tabular}{@{}lcc*{5}{c}@{\hspace{5pt}}*{5}{c}c@{}}
      \toprule
      & \multicolumn{2}{c}{}
        & \multicolumn{5}{c}{\textbf{RE10K}}
        & \multicolumn{5}{c}{\textbf{DL3DV}} & \\
      \cmidrule(lr){4-8}\cmidrule(lr){9-13}
      & \multicolumn{2}{c}{Input cameras} & \multicolumn{3}{c}{Appearance}
        & \multicolumn{2}{c}{Depth}
        & \multicolumn{3}{c}{Appearance}
        & \multicolumn{2}{c}{Depth} & \\
      \cmidrule(lr){2-3}\cmidrule(lr){4-6}\cmidrule(lr){7-8}
      \cmidrule(lr){9-11}\cmidrule(lr){12-13}
      \textbf{Method} & Intr. & Extr.
        & PSNR & SSIM & LPIPS & $\delta_{1.25}$ & AbsRel
        & PSNR & SSIM & LPIPS & $\delta_{1.25}$ & AbsRel
        & Runtime \\
      & & & $\uparrow$ & $\uparrow$ & $\downarrow$ & $\uparrow$ & $\downarrow$
        & $\uparrow$ & $\uparrow$ & $\downarrow$ & $\uparrow$ & $\downarrow$
        & (s)$\downarrow$ \\
      \midrule
      \multicolumn{14}{l}{\emph{Sparse-view} ($n=2$)} \\ [2pt]
      \quad pixelSplat & $\circ$ & $\circ$ & 26.125 & 0.864 & 0.131 & 0.608 & 0.359 & 23.358 & 0.721 & 0.239 & 0.489 & 0.467 & 0.088 \\
      \quad MVSplat & $\circ$ & $\circ$ & \cellcolor{tablesecond}26.319 & \cellcolor{tablesecond}0.865 & \cellcolor{tablesecond}0.127 & 0.769 & 0.220 & 22.894 & 0.704 & 0.235 & 0.687 & 0.275 & \cellcolor{tablebest}0.039 \\
      \quad DepthSplat & $\circ$ & $\circ$ & \cellcolor{tablebest}26.636 & \cellcolor{tablebest}0.881 & \cellcolor{tablebest}0.117 & 0.783 & 0.193 & \cellcolor{tablebest}24.078 & \cellcolor{tablesecond}0.759 & \cellcolor{tablebest}0.196 & 0.765 & 0.193 & 0.067 \\
      \quad VolSplat & $\circ$ & $\circ$ & 21.730 & 0.787 & 0.217 & 0.740 & 0.223 & 17.911 & 0.555 & 0.419 & 0.584 & 0.307 & 0.093 \\
      \quad PF3plat & $\circ$ & $\times$ & 20.011 & 0.638 & 0.261 & 0.806 & 0.173 & 17.743 & 0.500 & 0.348 & 0.710 & 0.231 & 0.327 \\
      \quad SPFSplat & $\circ$ & $\times$ & 21.850 & 0.707 & 0.178 & 0.814 & 0.174 & 20.892 & 0.610 & 0.238 & 0.799 & 0.170 & 0.053 \\
      \quad NoPoSplat & $\circ$ & $\times$ & 22.470 & 0.721 & 0.161 & 0.836 & 0.159 & 21.905 & 0.665 & 0.208 & 0.806 & 0.159 & \cellcolor{tablesecond}0.052 \\
      \quad Splatt3R & $\times$ & $\times$ & 20.731 & 0.688 & 0.272 & 0.835 & 0.131 & 19.635 & 0.622 & 0.320 & \cellcolor{tablebest}0.818 & \cellcolor{tablebest}0.140 & 0.811 \\
      \quad FreeSplatter & $\times$ & $\times$ & 22.973 & 0.739 & 0.206 & 0.809 & 0.165 & 22.097 & 0.642 & 0.271 & 0.734 & 0.222 & 0.201 \\
      \addlinespace[2pt]
      \quad \textbf{DensiTok} (AnySplat) & $\times$ & $\times$ & 24.628 & 0.830 & 0.158 & \cellcolor{tablesecond}0.864 & \cellcolor{tablesecond}0.127 & 22.111 & 0.711 & 0.218 & 0.798 & 0.167 & 0.071 \\
      \quad \textbf{DensiTok} (WorldMirror) & $\times$ & $\times$ & 25.079 & 0.838 & 0.162 & \cellcolor{tablebest}0.876 & \cellcolor{tablebest}0.108 & \cellcolor{tablesecond}24.056 & \cellcolor{tablebest}0.770 & \cellcolor{tablesecond}0.202 & \cellcolor{tablesecond}0.807 & \cellcolor{tablesecond}0.151 & 0.071 \\
      \quad \textbf{DensiTok} (DA3) & $\times$ & $\times$ & 24.233 & 0.818 & 0.166 & 0.846 & 0.137 & 21.816 & 0.702 & 0.226 & 0.780 & 0.180 & 0.081 \\
      \midrule
      \multicolumn{14}{l}{\emph{Sparse-view} ($n=3$)} \\ [2pt]
      \quad pixelSplat & $\circ$ & $\circ$ & \cellcolor{tablebest}28.184 & \cellcolor{tablebest}0.902 & \cellcolor{tablebest}0.102 & 0.692 & 0.287 & \cellcolor{tablesecond}25.721 & 0.813 & 0.167 & 0.564 & 0.366 & 0.138 \\
      \quad MVSplat & $\circ$ & $\circ$ & 24.644 & 0.866 & 0.136 & 0.658 & 0.329 & 22.184 & 0.728 & 0.228 & 0.586 & 0.400 & \cellcolor{tablebest}0.054 \\
      \quad DepthSplat & $\circ$ & $\circ$ & 26.399 & \cellcolor{tablesecond}0.901 & \cellcolor{tablesecond}0.105 & 0.786 & 0.182 & 24.425 & \cellcolor{tablesecond}0.819 & 0.155 & 0.793 & 0.166 & 0.092 \\
      \quad VolSplat & $\circ$ & $\circ$ & 25.532 & 0.871 & 0.144 & 0.782 & 0.193 & 21.960 & 0.676 & 0.322 & 0.715 & 0.223 & 0.131 \\
      \quad SelfSplat & $\circ$ & $\times$ & 22.001 & 0.732 & 0.231 & 0.723 & 0.225 & 16.440 & 0.446 & 0.478 & 0.607 & 0.279 & \cellcolor{tablesecond}0.057 \\
      \quad PF3plat & $\circ$ & $\times$ & 20.976 & 0.706 & 0.228 & 0.822 & 0.166 & 18.114 & 0.519 & 0.337 & 0.715 & 0.230 & 0.505 \\
      \quad SPFSplat & $\circ$ & $\times$ & 23.402 & 0.768 & 0.157 & 0.802 & 0.179 & 20.915 & 0.621 & 0.255 & 0.754 & 0.201 & 0.059 \\
      \quad NoPoSplat & $\circ$ & $\times$ & 22.298 & 0.702 & 0.152 & 0.833 & 0.158 & 20.889 & 0.588 & 0.228 & 0.804 & 0.165 & 0.060 \\
      \quad FreeSplatter & $\times$ & $\times$ & 20.401 & 0.639 & 0.308 & 0.767 & 0.184 & 20.147 & 0.560 & 0.349 & 0.702 & 0.240 & 0.238 \\
      \addlinespace[2pt]
      \quad \textbf{DensiTok} (AnySplat) & $\times$ & $\times$ & \cellcolor{tablesecond}26.919 & 0.873 & 0.124 & \cellcolor{tablesecond}0.866 & \cellcolor{tablesecond}0.123 & 24.948 & 0.800 & \cellcolor{tablesecond}0.154 & \cellcolor{tablesecond}0.824 & \cellcolor{tablesecond}0.148 & 0.086 \\
      \quad \textbf{DensiTok} (WorldMirror) & $\times$ & $\times$ & 26.818 & 0.869 & 0.137 & \cellcolor{tablebest}0.888 & \cellcolor{tablebest}0.100 & \cellcolor{tablebest}26.013 & \cellcolor{tablebest}0.830 & \cellcolor{tablebest}0.154 & \cellcolor{tablebest}0.832 & \cellcolor{tablebest}0.134 & 0.090 \\
      \quad \textbf{DensiTok} (DA3) & $\times$ & $\times$ & 26.634 & 0.864 & 0.133 & 0.848 & 0.136 & 24.716 & 0.793 & 0.161 & 0.811 & 0.157 & 0.098 \\
      \midrule
      \multicolumn{14}{l}{\emph{Sparse-view} ($n=5$)} \\ [2pt]
      \quad MVSplat & $\circ$ & $\circ$ & 22.219 & 0.836 & 0.166 & 0.517 & 0.497 & 20.741 & 0.724 & 0.243 & 0.489 & 0.528 & \cellcolor{tablesecond}0.093 \\
      \quad DepthSplat & $\circ$ & $\circ$ & 23.940 & 0.876 & 0.128 & 0.799 & 0.176 & 23.029 & 0.835 & 0.154 & 0.810 & 0.154 & 0.131 \\
      \quad VolSplat & $\circ$ & $\circ$ & 27.404 & \cellcolor{tablebest}0.903 & 0.117 & 0.798 & 0.183 & 24.254 & 0.760 & 0.248 & 0.790 & 0.174 & 0.188 \\
      \quad PF3plat & $\circ$ & $\times$ & 20.961 & 0.706 & 0.229 & 0.823 & 0.166 & 17.928 & 0.514 & 0.341 & 0.710 & 0.227 & 1.096 \\
      \quad SPFSplat & $\circ$ & $\times$ & 24.062 & 0.787 & 0.141 & 0.806 & 0.177 & 21.362 & 0.638 & 0.239 & 0.754 & 0.198 & \cellcolor{tablebest}0.072 \\
      \quad FreeSplatter & $\times$ & $\times$ & 19.644 & 0.611 & 0.354 & 0.760 & 0.198 & 19.627 & 0.538 & 0.384 & 0.697 & 0.249 & 0.487 \\
      \addlinespace[2pt]
      \quad \textbf{DensiTok} (AnySplat) & $\times$ & $\times$ & \cellcolor{tablebest}28.034 & \cellcolor{tablesecond}0.894 & \cellcolor{tablebest}0.108 & \cellcolor{tablesecond}0.866 & \cellcolor{tablesecond}0.121 & \cellcolor{tablesecond}26.795 & \cellcolor{tablesecond}0.844 & \cellcolor{tablebest}0.123 & \cellcolor{tablesecond}0.830 & \cellcolor{tablesecond}0.141 & 0.129 \\
      \quad \textbf{DensiTok} (WorldMirror) & $\times$ & $\times$ & 27.644 & 0.886 & 0.125 & \cellcolor{tablebest}0.894 & \cellcolor{tablebest}0.094 & \cellcolor{tablebest}27.154 & \cellcolor{tablebest}0.860 & \cellcolor{tablesecond}0.130 & \cellcolor{tablebest}0.842 & \cellcolor{tablebest}0.124 & 0.136 \\
      \quad \textbf{DensiTok} (DA3) & $\times$ & $\times$ & \cellcolor{tablesecond}27.713 & 0.886 & \cellcolor{tablesecond}0.114 & 0.847 & 0.132 & 26.540 & 0.836 & 0.130 & 0.818 & 0.149 & 0.145 \\
      \midrule
      \multicolumn{14}{l}{\emph{Sparse-view} ($n=7$)} \\ [2pt]
      \quad MVSplat & $\circ$ & $\circ$ & 21.244 & 0.820 & 0.181 & 0.530 & 0.488 & 20.080 & 0.717 & 0.252 & 0.497 & 0.508 & \cellcolor{tablesecond}0.147 \\
      \quad DepthSplat & $\circ$ & $\circ$ & 22.742 & 0.856 & 0.145 & 0.791 & 0.179 & 21.905 & 0.815 & 0.173 & 0.796 & 0.157 & 0.177 \\
      \quad VolSplat & $\circ$ & $\circ$ & \cellcolor{tablesecond}28.328 & \cellcolor{tablebest}0.909 & 0.113 & 0.801 & 0.179 & 25.007 & 0.783 & 0.227 & 0.799 & 0.163 & 0.256 \\
      \quad PF3plat & $\circ$ & $\times$ & 20.888 & 0.699 & 0.232 & 0.821 & 0.167 & 17.834 & 0.510 & 0.344 & 0.708 & 0.232 & 1.899 \\
      \quad SPFSplat & $\circ$ & $\times$ & 24.478 & 0.798 & 0.132 & 0.807 & 0.176 & 21.515 & 0.642 & 0.233 & 0.755 & 0.200 & \cellcolor{tablebest}0.096 \\
      \quad FreeSplatter & $\times$ & $\times$ & 19.203 & 0.596 & 0.378 & 0.748 & 0.210 & 19.173 & 0.521 & 0.408 & 0.680 & 0.257 & 0.760 \\
      \addlinespace[2pt]
      \quad \textbf{DensiTok} (AnySplat) & $\times$ & $\times$ & \cellcolor{tablebest}28.522 & \cellcolor{tablesecond}0.903 & \cellcolor{tablebest}0.101 & \cellcolor{tablesecond}0.865 & \cellcolor{tablesecond}0.120 & \cellcolor{tablesecond}27.627 & \cellcolor{tablesecond}0.865 & \cellcolor{tablebest}0.109 & \cellcolor{tablesecond}0.835 & \cellcolor{tablesecond}0.139 & 0.172 \\
      \quad \textbf{DensiTok} (WorldMirror) & $\times$ & $\times$ & 28.087 & 0.896 & 0.117 & \cellcolor{tablebest}0.898 & \cellcolor{tablebest}0.092 & \cellcolor{tablebest}27.693 & \cellcolor{tablebest}0.876 & 0.120 & \cellcolor{tablebest}0.850 & \cellcolor{tablebest}0.121 & 0.186 \\
      \quad \textbf{DensiTok} (DA3) & $\times$ & $\times$ & 28.225 & 0.895 & \cellcolor{tablesecond}0.108 & 0.851 & 0.129 & 27.463 & 0.861 & \cellcolor{tablesecond}0.114 & 0.821 & 0.147 & 0.193 \\
      \bottomrule
    \end{tabular}%
  }
  \endgroup
\end{table}

Figure~\ref{fig:app_gaussian_camera_visualization} compares the Gaussian scenes and camera poses predicted by AnySplat with and without DensiTok from the same sparse input views.
The Gaussian visualizations show that DensiTok reconstructs more coherent scene structure with clearer object boundaries.

The camera visualizations show improvements in both camera positions and viewing orientations.
In the illustrated examples, DensiTok's predicted camera centers lie closer to the ground-truth centers, and the predicted camera frusta align more closely with their ground-truth counterparts.
The reduced positional displacement and angular mismatch are particularly visible in the outdoor example.
These observations are consistent with the camera AUC improvements reported in Table~\ref{tab:main_results}.

The joint improvement in scene reconstruction and camera estimation reflects their dependence on a shared geometry representation.
As described in Section~\ref{sec:app_appfusion}, the pretrained camera head decodes the final camera poses from the completed geometry tokens.
The predicted and interpolated cameras in Section~\ref{sec:camera_conditioning} provide conditions for latent generation, while the final output cameras are predicted again from the enriched tokens.
Geometry token completion therefore benefits both Gaussian reconstruction and camera localization and orientation, even though the pretrained camera head remains frozen.

\begin{figure}[t]
  \centering
  \includegraphics[width=\textwidth]{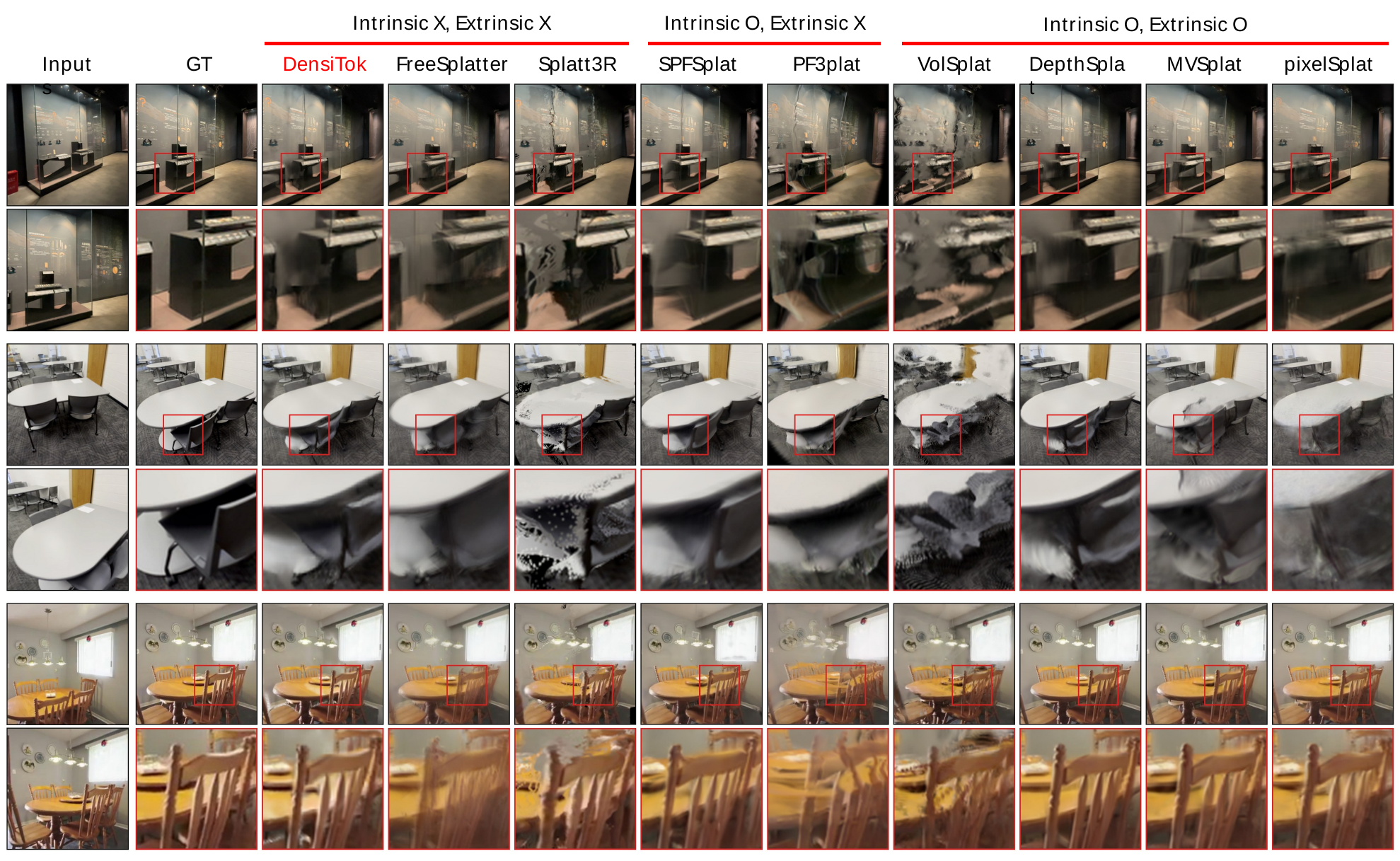}
  \caption{Qualitative novel-view RGB comparisons with sparse-view feed-forward Gaussian Splatting models. DensiTok uses the AnySplat backbone and requires neither input camera intrinsics nor extrinsics. Methods are grouped by the input camera information they require. The leftmost column shows two input views for each scene, and GT denotes the ground-truth target image. Red boxes mark regions enlarged below each rendering.}
  \label{fig:qualitative_results_other_sparse_methods}
\end{figure}

\subsection{Comparison with Sparse-View Feed-Forward Gaussian Splatting Models}
\label{sec:app_sparse_comparison}

\paragraph{Evaluation protocol.}
We compare DensiTok with existing feed-forward Gaussian Splatting models designed for sparse-view reconstruction, including pixelSplat~\citep{charatan2024pixelsplat}, MVSplat~\citep{chen2024mvsplat}, DepthSplat~\citep{xu2025depthsplat}, VolSplat~\citep{wang2025volsplat}, SelfSplat~\citep{kang2025selfsplat}, PF3plat~\citep{hong2025pf3plat}, SPFSplat~\citep{huang2025no}, NoPoSplat~\citep{ye2025no}, Splatt3R~\citep{smart2024splatt3r}, and FreeSplatter~\citep{xu2025freesplatter} on RE10K and DL3DV. Most existing sparse-view methods have been evaluated at $256\times256$ resolution~\citep{charatan2024pixelsplat, chen2024mvsplat, xu2025depthsplat, hong2025pf3plat}, so we evaluate all methods, including DensiTok, at this resolution. For DensiTok, which requires input image dimensions divisible by $14$, we perform inference at $252\times252$ and resize the predictions to the $256\times256$ evaluation resolution. Because this resolution differs from that of our original evaluation, we perform test-view sampling again. The resulting sampling indices differ from those used in Tables~\ref{tab:main_results} and~\ref{tab:app_more_quantitative}.

We consider $n\in\{2,3,5,7\}$ and evaluate each model only at input image counts supported by its official implementation. Table~\ref{tab:app_sparse_comparison} lists the evaluated method and input-count combinations. Unlisted combinations were not evaluated. The table reports PSNR, SSIM, LPIPS, depth accuracy $\delta_{1.25}$, AbsRel, and runtime, using the metric notation in Section~\ref{sec:exp_setup}.

\begin{table}[t]
  \centering
  \caption{Comparison with diffusion-based novel-view synthesis methods on RE10K and DL3DV at $512\times512$ with $n\in\{2,3,5,7\}$. Runtime is reported in seconds in a shared column. \protect\colorbox{tablebest}{Orange} and \protect\colorbox{tablesecond}{yellow} mark the best and second-best values for each metric at the same $n$, respectively. Rankings are determined before rounding, with ties sharing the same color.}
  \label{tab:app_nvs_comparison}
  \begingroup
  \scriptsize
  \setlength{\tabcolsep}{2pt}
  \renewcommand{\arraystretch}{1.15}
  \resizebox{0.75\textwidth}{!}{%
    \begin{tabular}{@{}l*{3}{c}@{\hspace{5pt}}*{3}{c}c@{}}
      \toprule
      & \multicolumn{3}{c}{\textbf{RE10K}}
        & \multicolumn{3}{c}{\textbf{DL3DV}} & \\
      \cmidrule(lr){2-4}\cmidrule(lr){5-7}
      \textbf{Method}
        & PSNR & SSIM & LPIPS
        & PSNR & SSIM & LPIPS
        & Runtime \\
      & $\uparrow$ & $\uparrow$ & $\downarrow$
        & $\uparrow$ & $\uparrow$ & $\downarrow$
        & (s)$\downarrow$ \\
      \midrule
      \multicolumn{8}{l}{\emph{Sparse-view} ($n=2$)} \\ [2pt]
      \quad NVComposer & 14.897 & 0.528 & 0.508 & 12.078 & 0.359 & 0.610 & 48.502 \\
      \quad Matrix3D & 20.934 & 0.700 & 0.284 & 18.149 & 0.523 & 0.376 & 25.592 \\
      \quad GLD & 23.608 & 0.768 & 0.219 & 21.440 & 0.621 & \cellcolor{tablesecond}0.270 & 76.824 \\
      \addlinespace[2pt]
      \quad \textbf{DensiTok} (AnySplat) & 24.435 & 0.825 & \cellcolor{tablebest}0.208 & 21.500 & 0.685 & \cellcolor{tablebest}0.270 & \cellcolor{tablebest}0.510 \\
      \quad \textbf{DensiTok} (WorldMirror) & \cellcolor{tablebest}24.882 & \cellcolor{tablebest}0.834 & \cellcolor{tablesecond}0.213 & \cellcolor{tablebest}23.448 & \cellcolor{tablebest}0.744 & 0.275 & \cellcolor{tablesecond}0.515 \\
      \quad \textbf{DensiTok} (DA3) & \cellcolor{tablesecond}24.598 & \cellcolor{tablesecond}0.827 & 0.226 & \cellcolor{tablesecond}22.503 & \cellcolor{tablesecond}0.709 & 0.285 & 0.579 \\
      \midrule
      \multicolumn{8}{l}{\emph{Sparse-view} ($n=3$)} \\ [2pt]
      \quad NVComposer & 15.518 & 0.545 & 0.482 & 12.254 & 0.362 & 0.600 & 48.486 \\
      \quad Matrix3D & 21.996 & 0.725 & 0.259 & 18.954 & 0.545 & 0.347 & 29.415 \\
      \quad GLD & 25.130 & 0.796 & 0.192 & 22.771 & 0.656 & 0.233 & 90.549 \\
      \addlinespace[2pt]
      \quad \textbf{DensiTok} (AnySplat) & \cellcolor{tablesecond}26.826 & \cellcolor{tablesecond}0.867 & \cellcolor{tablebest}0.170 & 24.212 & 0.764 & \cellcolor{tablebest}0.206 & \cellcolor{tablebest}0.531 \\
      \quad \textbf{DensiTok} (WorldMirror) & 26.720 & 0.863 & 0.183 & \cellcolor{tablebest}25.276 & \cellcolor{tablebest}0.793 & \cellcolor{tablesecond}0.211 & \cellcolor{tablesecond}0.553 \\
      \quad \textbf{DensiTok} (DA3) & \cellcolor{tablebest}26.985 & \cellcolor{tablebest}0.872 & \cellcolor{tablesecond}0.180 & \cellcolor{tablesecond}24.452 & \cellcolor{tablesecond}0.776 & 0.222 & 0.605 \\
      \midrule
      \multicolumn{8}{l}{\emph{Sparse-view} ($n=5$)} \\ [2pt]
      \quad Matrix3D & 22.684 & 0.738 & 0.246 & 19.480 & 0.557 & 0.333 & 51.567 \\
      \quad GLD & 26.100 & 0.813 & 0.176 & 23.446 & 0.677 & 0.214 & 118.995 \\
      \addlinespace[2pt]
      \quad \textbf{DensiTok} (AnySplat) & \cellcolor{tablesecond}27.995 & \cellcolor{tablesecond}0.888 & \cellcolor{tablebest}0.152 & \cellcolor{tablesecond}26.041 & 0.807 & \cellcolor{tablebest}0.172 & \cellcolor{tablebest}0.593 \\
      \quad \textbf{DensiTok} (WorldMirror) & 27.610 & 0.880 & 0.168 & \cellcolor{tablebest}26.396 & \cellcolor{tablebest}0.822 & \cellcolor{tablesecond}0.179 & \cellcolor{tablesecond}0.627 \\
      \quad \textbf{DensiTok} (DA3) & \cellcolor{tablebest}28.133 & \cellcolor{tablebest}0.891 & \cellcolor{tablesecond}0.163 & 25.717 & \cellcolor{tablesecond}0.812 & 0.192 & 0.668 \\
      \midrule
      \multicolumn{8}{l}{\emph{Sparse-view} ($n=7$)} \\ [2pt]
      \quad Matrix3D & 22.993 & 0.750 & 0.236 & 19.854 & 0.569 & 0.318 & 110.844 \\
      \quad GLD & 26.918 & 0.834 & 0.156 & 23.919 & 0.689 & 0.203 & 150.433 \\
      \addlinespace[2pt]
      \quad \textbf{DensiTok} (AnySplat) & \cellcolor{tablesecond}28.527 & \cellcolor{tablesecond}0.897 & \cellcolor{tablebest}0.143 & \cellcolor{tablesecond}26.841 & 0.826 & \cellcolor{tablebest}0.157 & \cellcolor{tablebest}0.656 \\
      \quad \textbf{DensiTok} (WorldMirror) & 28.087 & 0.890 & 0.153 & \cellcolor{tablebest}26.903 & \cellcolor{tablebest}0.837 & \cellcolor{tablesecond}0.168 & \cellcolor{tablesecond}0.713 \\
      \quad \textbf{DensiTok} (DA3) & \cellcolor{tablebest}28.762 & \cellcolor{tablebest}0.905 & \cellcolor{tablesecond}0.149 & 26.544 & \cellcolor{tablesecond}0.833 & 0.174 & 0.740 \\
      \bottomrule
    \end{tabular}%
  }
  \endgroup
\end{table}

\paragraph{Camera inputs.}
As shown in the Table~\ref{tab:app_sparse_comparison}, the Intr. and Extr. columns distinguish externally supplied input camera intrinsics and extrinsics required for reconstruction. These labels describe the evaluated configurations and exclude target cameras used only for rendering and evaluation. pixelSplat, MVSplat, and DepthSplat use both intrinsics and extrinsics. SelfSplat, PF3plat, and the evaluated SPFSplat configuration use intrinsics but estimate poses. NoPoSplat also uses input intrinsics but does not require input extrinsics. Splatt3R, FreeSplatter, and all three DensiTok variants require neither input intrinsics nor input extrinsics. This distinction makes the different camera information available to the compared methods explicit.

\paragraph{Comparison results.}
Most sparse-view methods report results using only two or three input views, while our protocol keeps the sampled evaluation views fixed across different input view counts. This difference in evaluation sampling can lead to slight discrepancies with the results reported in the original papers.

Table~\ref{tab:app_sparse_comparison} shows that all three DensiTok variants outperform FreeSplatter in PSNR, SSIM, and LPIPS on both datasets at all evaluated input counts, except for the DA3 variant in PSNR on DL3DV at $n=2$. None of the DensiTok variants requires input camera intrinsics or extrinsics. At $n=3$, all three variants also outperform NoPoSplat in every reported appearance and depth metric on both datasets. Among all evaluated methods, DepthSplat achieves the highest PSNR and lowest LPIPS on both datasets at $n=2$. DensiTok with AnySplat achieves the highest PSNR and lowest LPIPS on RE10K at $n\in\{5,7\}$, while DensiTok with WorldMirror achieves the highest PSNR and SSIM on DL3DV at $n\in\{3,5,7\}$. The DA3 variant achieves the second-highest PSNR on RE10K at $n=5$ and the second-lowest LPIPS on RE10K at $n\in\{5,7\}$ and on DL3DV at $n=7$.

DensiTok with WorldMirror also achieves the best values for both depth metrics on RE10K at every evaluated input count and on DL3DV at $n\in\{3,5,7\}$. Splatt3R remains best on both DL3DV depth metrics at $n=2$. All three DensiTok variants outperform FreeSplatter in both depth metrics on both datasets at $n\in\{2,3,5,7\}$.

The qualitative comparison in Figure~\ref{fig:qualitative_results_other_sparse_methods} further illustrates the reconstruction quality of DensiTok with the AnySplat backbone. Among methods that require neither input camera intrinsics nor extrinsics, DensiTok produces clearer object boundaries and fewer structural artifacts than FreeSplatter and Splatt3R in the illustrated examples. DensiTok also achieves better or comparable visual quality relative to the methods that require input intrinsics, with or without extrinsics. These examples show that DensiTok can preserve fine scene details without relying on externally supplied input camera information.

\begin{figure}[t]
  \centering
  \includegraphics[width=\textwidth]{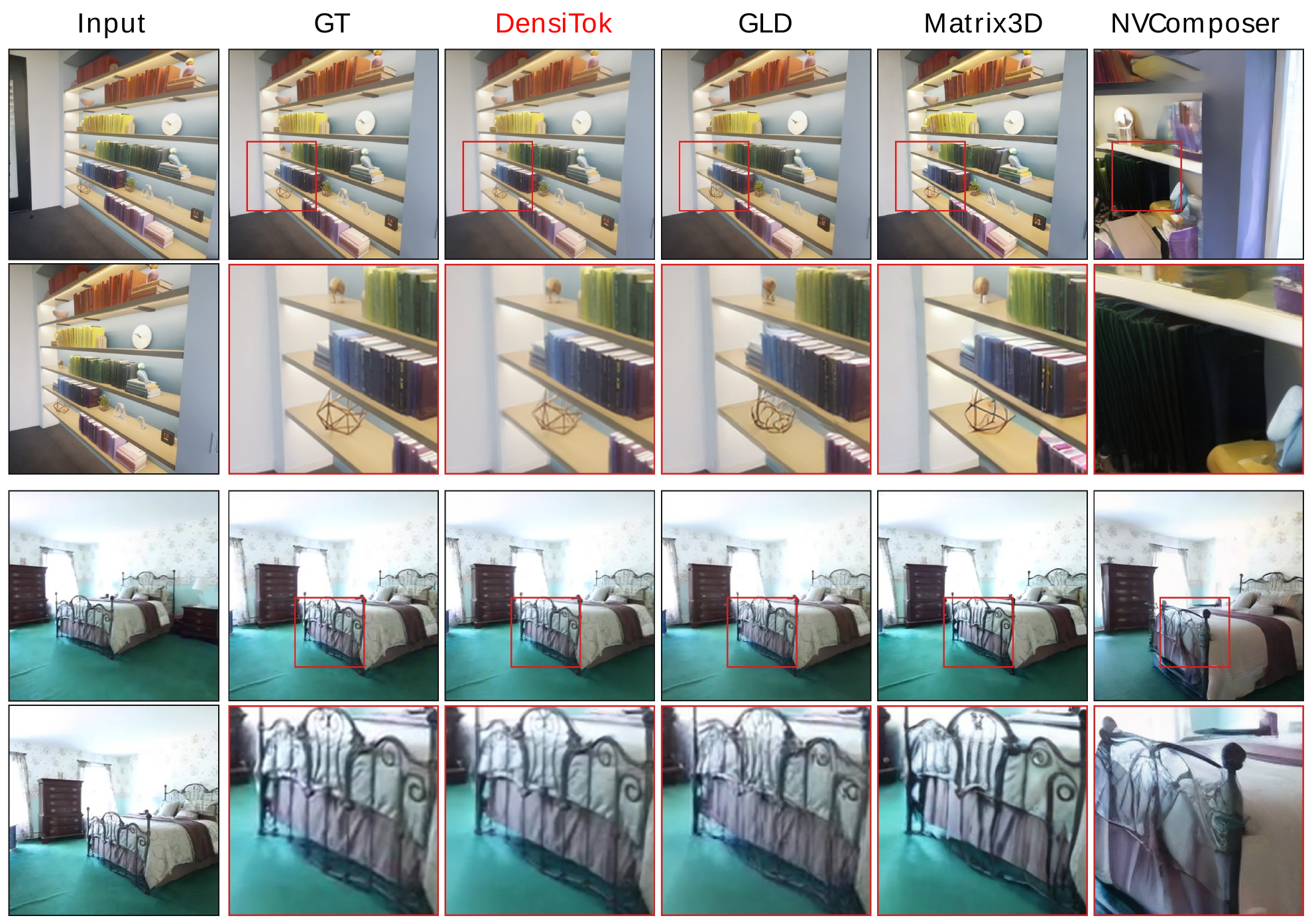}
  \caption{Qualitative novel-view RGB comparisons of DensiTok with GLD, Matrix3D, and NVComposer using $n=2$ input views. The leftmost column shows the input views for each scene, and GT denotes the ground-truth target image. Red boxes mark regions enlarged below each rendering to compare the preservation of scene content and local geometry.}
  \label{fig:qualitative_results_other_nvs_methods}
\end{figure}

\subsection{Comparison with Diffusion-Based Novel View Synthesis Models}
\label{sec:app_nvs_comparison}

\paragraph{Evaluation protocol.}
We compare DensiTok with NVComposer~\citep{li2025nvcomposer}, Matrix3D~\citep{lu2025matrix3d}, and GLD~\citep{jang2026repurposing} on RE10K and DL3DV. These methods use diffusion for novel-view image generation, with GLD operating in geometric feature space. All methods are evaluated at $512\times512$ resolution. GLD and DensiTok require input image dimensions divisible by $14$, so we perform inference at $504\times504$ and resize the predicted images to $512\times512$ before evaluation, following the resizing procedure in the preceding comparison. Table~\ref{tab:app_nvs_comparison} reports PSNR, SSIM, LPIPS, and runtime for the evaluated method and input-count combinations with $n\in\{2,3,5,7\}$.

\begin{table}[t]
  \begin{center}
    \caption{Generalization to ETH3D, ScanNet, and Mip-NeRF 360 with the AnySplat backbone and $n\in\{2,3,5,7\}$ observed views. Camera AUC is reported in percent. Depth is evaluated against official ground truth on ETH3D and ScanNet and MoGe-2 predictions on Mip-NeRF 360.}
    \resizebox{0.85\columnwidth}{!}{%
      \centering
      \setlength{\tabcolsep}{2pt}
      \renewcommand{\arraystretch}{1.2}
      \begin{tabular}{cl*{7}{c}}
        \toprule
        \multirow{2}{*}{$n$} & \multirow{2}{*}{\quad Methods}
          & \multicolumn{3}{c}{Appearance} & \multicolumn{2}{c}{Camera} & \multicolumn{2}{c}{Depth} \\
        \cmidrule(lr){3-5}\cmidrule(lr){6-7}\cmidrule(lr){8-9}
          & & PSNR$\uparrow$ & SSIM$\uparrow$ & LPIPS$\downarrow$ & AUC@$3^{\circ}\uparrow$ & AUC@$30^{\circ}\uparrow$ & $\delta_{1.25}\uparrow$ & AbsRel$\downarrow$ \\
        \midrule
        \multicolumn{9}{l}{\emph{ETH3D}~\citep{schops2017multi}} \\ [2pt]
        \multirow{2}{*}{$2$} & \quad AnySplat & 15.99 & 0.462 & 0.489 & 27.3 & 89.4 & 0.767 & 0.148 \\
         & \quad\textbf{+ DensiTok} & \textbf{17.01} & \textbf{0.531} & \textbf{0.466} & \textbf{33.3} & \textbf{90.3} & \textbf{0.771} & \textbf{0.135} \\ [2pt]
        \multirow{2}{*}{$3$} & \quad AnySplat & 16.79 & 0.487 & 0.461 & 29.3 & 88.0 & 0.831 & 0.119 \\
         & \quad\textbf{+ DensiTok} & \textbf{17.92} & \textbf{0.554} & \textbf{0.447} & \textbf{31.3} & \textbf{89.2} & \textbf{0.837} & \textbf{0.116} \\ [2pt]
        \multirow{2}{*}{$5$} & \quad AnySplat & 16.53 & 0.481 & 0.473 & 41.2 & 90.6 & 0.859 & 0.114 \\
         & \quad\textbf{+ DensiTok} & \textbf{17.99} & \textbf{0.565} & \textbf{0.437} & \textbf{41.8} & \textbf{90.8} & \textbf{0.869} & \textbf{0.099} \\ [2pt]
        \multirow{2}{*}{$7$} & \quad AnySplat & 16.90 & 0.487 & 0.467 & 39.5 & 89.7 & 0.874 & 0.101 \\
         & \quad\textbf{+ DensiTok} & \textbf{18.09} & \textbf{0.580} & \textbf{0.415} & \textbf{41.6} & \textbf{91.1} & \textbf{0.889} & \textbf{0.092} \\
        \midrule
        \multicolumn{9}{l}{\emph{ScanNet}~\citep{dai2017scannet}} \\ [2pt]
        \multirow{2}{*}{$2$} & \quad AnySplat & 19.60 & 0.668 & 0.374 & 13.8 & 77.5 & 0.952 & 0.070 \\
         & \quad\textbf{+ DensiTok} & \textbf{22.51} & \textbf{0.761} & \textbf{0.319} & \textbf{14.7} & \textbf{77.6} & \textbf{0.972} & \textbf{0.052} \\ [2pt]
        \multirow{2}{*}{$3$} & \quad AnySplat & 21.63 & 0.714 & 0.332 & 13.9 & 76.6 & 0.973 & 0.057 \\
         & \quad\textbf{+ DensiTok} & \textbf{24.18} & \textbf{0.803} & \textbf{0.278} & \textbf{16.3} & \textbf{78.2} & \textbf{0.981} & \textbf{0.048} \\ [2pt]
        \multirow{2}{*}{$5$} & \quad AnySplat & 22.68 & 0.740 & 0.314 & 15.4 & 75.6 & 0.977 & 0.053 \\
         & \quad\textbf{+ DensiTok} & \textbf{24.76} & \textbf{0.822} & \textbf{0.264} & \textbf{17.1} & \textbf{76.4} & \textbf{0.981} & \textbf{0.047} \\ [2pt]
        \multirow{2}{*}{$7$} & \quad AnySplat & 23.25 & 0.755 & 0.304 & 14.5 & 74.3 & 0.979 & 0.052 \\
         & \quad\textbf{+ DensiTok} & \textbf{25.20} & \textbf{0.833} & \textbf{0.252} & \textbf{15.5} & \textbf{74.9} & \textbf{0.982} & \textbf{0.046} \\
        \midrule
        \multicolumn{9}{l}{\emph{Mip-NeRF 360}~\citep{barron2022mip}} \\ [2pt]
        \multirow{2}{*}{$2$} & \quad AnySplat & 17.58 & 0.428 & 0.395 & \textbf{14.8} & 73.7 & 0.779 & 0.168 \\
         & \quad\textbf{+ DensiTok} & \textbf{19.27} & \textbf{0.490} & \textbf{0.387} & \textbf{14.8} & \textbf{82.6} & \textbf{0.814} & \textbf{0.155} \\ [2pt]
        \multirow{2}{*}{$3$} & \quad AnySplat & 18.05 & 0.450 & \textbf{0.383} & 11.1 & 78.6 & 0.796 & 0.168 \\
         & \quad\textbf{+ DensiTok} & \textbf{19.24} & \textbf{0.469} & 0.400 & \textbf{12.3} & \textbf{83.6} & \textbf{0.812} & \textbf{0.159} \\ [2pt]
        \multirow{2}{*}{$5$} & \quad AnySplat & 18.73 & 0.470 & 0.383 & 11.5 & 82.1 & 0.794 & 0.166 \\
         & \quad\textbf{+ DensiTok} & \textbf{20.02} & \textbf{0.498} & \textbf{0.371} & \textbf{13.0} & \textbf{83.6} & \textbf{0.829} & \textbf{0.153} \\ [2pt]
        \multirow{2}{*}{$7$} & \quad AnySplat & 19.76 & 0.501 & 0.358 & 15.5 & 84.8 & 0.809 & 0.160 \\
         & \quad\textbf{+ DensiTok} & \textbf{20.48} & \textbf{0.521} & \textbf{0.349} & \textbf{16.0} & \textbf{86.3} & \textbf{0.834} & \textbf{0.151} \\
        \bottomrule
      \end{tabular}
    }
    \label{tab:app_generalization}
  \end{center}
\end{table}

\paragraph{Comparison results.}
Most diffusion-based NVS methods aim to synthesize plausible content in unobserved regions and are typically evaluated with widely spaced input views. Our reconstruction task requires a minimum degree of overlap between input views, so we use more closely spaced inputs. This difference in view sampling can lead to modest discrepancies with the results reported in the original papers.

Table~\ref{tab:app_nvs_comparison} shows that all three DensiTok variants achieve higher PSNR and SSIM than the evaluated diffusion baselines on both datasets at every matching input count. DensiTok with DA3 achieves the highest PSNR and SSIM on RE10K for $n\in\{3,5,7\}$, while the WorldMirror variant leads these metrics on DL3DV at every evaluated input count. The AnySplat variant achieves the lowest LPIPS on both datasets across all input counts. All three variants also outperform the diffusion baselines in LPIPS for $n\in\{3,5,7\}$. At $n=2$, GLD retains lower LPIPS than the DA3 variant on both datasets and the WorldMirror variant on DL3DV. The qualitative comparison in Figure~\ref{fig:qualitative_results_other_nvs_methods} shows that the evaluated diffusion baselines can alter scene context and exhibit weaker geometric consistency with the observed scene. NVComposer changes the bookshelf composition and bed appearance, producing plausible content that departs from the target scene. GLD and Matrix3D better retain the overall layout, but the enlarged shelf ornament and bed-frame regions reveal local shape distortions. In these examples, DensiTok more closely preserves the scene content, object arrangement, and geometric structure of the ground-truth targets. DensiTok completes inference in under a second, while the evaluated diffusion baselines require tens of seconds to a few minutes. This yields speedups of tens to hundreds of times at matching input counts. At $n=2$, DensiTok requires 0.510 to 0.579 seconds compared with 76.824 seconds for GLD, yielding approximately $133\times$ to $151\times$ speedups across the three backbones. DensiTok therefore combines stronger image reconstruction metrics with substantially lower latency, making it more practical for fast novel-view synthesis.

\section{Additional Ablation Studies}
\label{sec:app_additional_ablations}

In this section, we use AnySplat as a common reconstruction backbone for all experiments, following the ablation studies in the main paper.

\subsection{Generalization to Unseen Datasets}
\label{sec:app_generalization}

\paragraph{Evaluation protocol.}
We evaluate generalization on ETH3D~\citep{schops2017multi}, ScanNet~\citep{dai2017scannet}, and Mip-NeRF 360~\citep{barron2022mip}, none of which is used to train DensiTok. We compare AnySplat with and without DensiTok using $n\in\{2,3,5,7\}$ observed views.
We use the official evaluation sequences for each dataset, comprising 11 sequences for ETH3D, 75 for ScanNet, and 9 for Mip-NeRF 360.
We evaluate four novel views per sequence on ScanNet.
For ETH3D and Mip-NeRF 360, we reduce the number of evaluation views to one per sequence because fewer views are available in each sequence.

For depth evaluation, we use the official ground-truth depth provided by ETH3D and ScanNet. Since Mip-NeRF 360 does not provide ground-truth depth, we use MoGe-2~\citep{wang2025moge2} predictions as pseudo-labels, following the main evaluation.

\paragraph{Generalization results.}
Table~\ref{tab:app_generalization} shows consistent improvements across the three datasets, with PSNR typically increasing by around one to three decibels. PSNR gains are largest on ScanNet, while ETH3D and Mip-NeRF 360 also benefit across input counts. Camera and depth accuracy improve throughout, indicating that the gains extend beyond image reconstruction. These results support the ability of geometry token completion to improve reconstruction on datasets excluded from DensiTok training.

\begin{table}[t]
  \centering
  \caption{Effect of one-step finetuning with AnySplat, $d_z=4$, $n=2$, and $N=16$. FM denotes flow matching and NFE denotes the number of DiT forward evaluations. Runtime is reported in a shared column.}
  \label{tab:onestep_finetuning}
  \begingroup
  \scriptsize
  \setlength{\tabcolsep}{3pt}
  \renewcommand{\arraystretch}{1.15}
  \resizebox{0.85\textwidth}{!}{%
    \begin{tabular}{@{}lc*{3}{c}@{\hspace{6pt}}*{3}{c}@{\hspace{6pt}}c@{}}
      \toprule
      & & \multicolumn{3}{c}{\textbf{RE10K}}
        & \multicolumn{3}{c}{\textbf{DL3DV}}
        & \textbf{Efficiency} \\
      \cmidrule(lr){3-5}\cmidrule(lr){6-8}\cmidrule(lr){9-9}
      \textbf{Method} & NFE
        & PSNR & SSIM & LPIPS
        & PSNR & SSIM & LPIPS
        & Runtime \\
      & $\downarrow$
        & $\uparrow$ & $\uparrow$ & $\downarrow$
        & $\uparrow$ & $\uparrow$ & $\downarrow$
        & (s)$\downarrow$ \\
      \midrule
      Deterministic regression & 1 & 20.86 & 0.712 & 0.246 & 19.97 & 0.631 & 0.288 & 0.258 \\
      \midrule
      One-step training from scratch & 1 & 21.67 & 0.746 & 0.240 & 21.28 & 0.670 & 0.285 & 0.258 \\
      \midrule
      \multirow{3}{*}{FM pretraining only}
        & 1 & 21.56 & 0.743 & 0.243 & 21.09 & 0.665 & 0.289 & 0.258 \\
        & 8 & 21.84 & 0.755 & 0.236 & 21.59 & 0.676 & 0.279 & 0.990 \\
        & 50 & 22.86 & 0.773 & 0.217 & 22.81 & 0.709 & 0.246 & 4.848 \\
      \midrule
      \textbf{FM + one-step finetuning} & 1 & 22.56 & 0.766 & 0.223 & 22.40 & 0.699 & 0.255 & 0.258 \\
      \bottomrule
    \end{tabular}%
  }
  \endgroup
\end{table}

\subsection{Effect of flow-matching pretraining and one-step finetuning.}
\label{sec:app_flow_matching_finetuning}

We evaluate whether one-step finetuning preserves the reconstruction quality of iterative sampling while reducing inference cost. We also compare against deterministic regression and one-step training from scratch to examine the benefit of flow-matching pretraining. Table~\ref{tab:onestep_finetuning} reports these results with the AnySplat backbone, $d_z=4$, $n=2$, and $N=16$, including the Stage~2 checkpoint evaluated at different NFEs and its finetuned counterpart.

Both baselines retain the camera-conditioned DiT architecture, pretrained backbone, and Stage~1 LAM-VAE, but train the DiT from random initialization without flow-matching pretraining. Let $c$ denote the observed latents $\mathbf{Z}^{\mathrm{obs}}$, view mask $\mathbf{m}$, and cameras $\{\mathbf{g}_i\}_{i=1}^N$. \emph{Deterministic regression} uses zero inputs for unobserved views and directly predicts $\tilde{\mathbf{z}}_i=f_{\theta,i}(\mathbf{0},c)$. \emph{One-step training from scratch} instead fixes the time of the unobserved views at $t=1$ and predicts $\tilde{\mathbf{z}}_i=\boldsymbol{\epsilon}_i-\widehat{\mathbf{v}}_{\theta,i}$ with $\boldsymbol{\epsilon}_i\sim\mathcal{N}(\mathbf{0},\mathbf{I})$, following the Stage~3 parameterization. Both baselines use the same loss terms as in Stage~3.

One-step finetuning retains most of the rendering quality of the 50-evaluation reference while achieving a speedup of approximately $19\times$. It also outperforms the unfinetuned model at one or eight evaluations. At the same single-evaluation cost, it achieves better PSNR, SSIM, and LPIPS than both deterministic regression and one-step training from scratch on RE10K and DL3DV. These results support combining flow-matching pretraining with one-step finetuning for efficient latent completion.

\subsection{Effect of Latent Channel Count}
\label{sec:app_latent_channels}

Table~\ref{tab:latent_channels} compares DensiTok with $d_z=4$ and $d_z=16$ latent channels on RE10K and DL3DV after one-step finetuning. Both variants use the same evaluation views, respectively.

Increasing the latent channel count from $4$ to $16$ expands the representational capacity of the latent space and relaxes the compression bottleneck. The 16-channel variant improves most appearance, camera, and depth metrics across both datasets and input counts. These results suggest that a wider latent representation can better preserve appearance and geometric information, although improvements are not uniform across all metrics.

\begin{table}[t]
  \centering
  \caption{Effect of latent channel count with AnySplat, $N=16$, and one-step inference.}
  \label{tab:latent_channels}
  \begingroup
  \scriptsize
  \setlength{\tabcolsep}{2pt}
  \renewcommand{\arraystretch}{1.15}
  \resizebox{\textwidth}{!}{%
    \begin{tabular}{@{}cc*{7}{c}@{\hspace{6pt}}*{7}{c}@{}}
      \toprule
      &
        & \multicolumn{7}{c}{\textbf{RE10K}}
        & \multicolumn{7}{c}{\textbf{DL3DV}} \\
      \cmidrule(lr){3-9}\cmidrule(lr){10-16}
        & & \multicolumn{3}{c}{Appearance} & \multicolumn{2}{c}{Camera} & \multicolumn{2}{c}{Depth}
        & \multicolumn{3}{c}{Appearance} & \multicolumn{2}{c}{Camera} & \multicolumn{2}{c}{Depth} \\
      \cmidrule(lr){3-5}\cmidrule(lr){6-7}\cmidrule(lr){8-9}
      \cmidrule(lr){10-12}\cmidrule(lr){13-14}\cmidrule(lr){15-16}
      $n$ & $d_z$ & PSNR$\uparrow$ & SSIM$\uparrow$ & LPIPS$\downarrow$ & AUC@$3^{\circ}\uparrow$ & AUC@$30^{\circ}\uparrow$ & $\delta_{1.25}\uparrow$ & AbsRel$\downarrow$
        & PSNR$\uparrow$ & SSIM$\uparrow$ & LPIPS$\downarrow$ & AUC@$3^{\circ}\uparrow$ & AUC@$30^{\circ}\uparrow$ & $\delta_{1.25}\uparrow$ & AbsRel$\downarrow$ \\
      \midrule
      \multirow{2}{*}{$2$} & $4$ & 22.56 & 0.766 & 0.223 & 26.7 & 78.9 & 0.846 & 0.153 & \textbf{22.40} & 0.699 & 0.255 & 30.4 & 88.9 & 0.801 & 0.170 \\
       & $16$ & \textbf{22.64} & \textbf{0.770} & \textbf{0.215} & \textbf{27.5} & \textbf{79.0} & \textbf{0.854} & \textbf{0.151} & 22.24 & \textbf{0.700} & \textbf{0.247} & \textbf{39.6} & \textbf{90.8} & \textbf{0.804} & \textbf{0.164} \\
      \midrule
      \multirow{2}{*}{$3$} & $4$ & 24.28 & 0.813 & 0.184 & 25.8 & \textbf{79.7} & 0.853 & 0.146 & 24.28 & 0.755 & 0.205 & 34.4 & 89.6 & 0.809 & 0.158 \\
       & $16$ & \textbf{24.57} & \textbf{0.822} & \textbf{0.171} & \textbf{27.1} & 79.5 & \textbf{0.862} & \textbf{0.145} & \textbf{24.29} & \textbf{0.760} & \textbf{0.194} & \textbf{38.5} & \textbf{90.5} & \textbf{0.812} & \textbf{0.155} \\
      \midrule
      \multirow{2}{*}{$5$} & $4$ & 25.35 & 0.838 & 0.162 & 25.2 & 78.8 & 0.852 & 0.141 & 25.33 & 0.782 & 0.182 & 33.5 & 87.7 & \textbf{0.816} & 0.153 \\
       & $16$ & \textbf{25.88} & \textbf{0.849} & \textbf{0.147} & \textbf{25.9} & \textbf{79.1} & \textbf{0.860} & \textbf{0.132} & \textbf{25.56} & \textbf{0.794} & \textbf{0.168} & \textbf{40.0} & \textbf{89.6} & \textbf{0.816} & \textbf{0.151} \\
      \midrule
      \multirow{2}{*}{$7$} & $4$ & 25.89 & 0.851 & 0.150 & 24.5 & 78.5 & 0.853 & 0.137 & 25.71 & 0.796 & 0.169 & 34.0 & 86.7 & \textbf{0.817} & 0.151 \\
       & $16$ & \textbf{26.45} & \textbf{0.862} & \textbf{0.138} & \textbf{25.6} & \textbf{79.1} & \textbf{0.857} & \textbf{0.130} & \textbf{26.04} & \textbf{0.810} & \textbf{0.158} & \textbf{39.0} & \textbf{88.8} & 0.816 & \textbf{0.150} \\
      \bottomrule
    \end{tabular}%
  }
  \endgroup
\end{table}

\begin{table}[t]
  \centering
  \caption{Effect of Gaussian head finetuning with AnySplat, $d_z=4$, $N=16$, and one-step inference. Image embedding only denotes finetuning $h_{\mathrm{img}}$ while keeping the remaining pretrained Gaussian head weights frozen.}
  \label{tab:gaussian_head_finetuning}
  \begingroup
  \scriptsize
  \setlength{\tabcolsep}{2pt}
  \renewcommand{\arraystretch}{1.15}
  \resizebox{\textwidth}{!}{%
    \begin{tabular}{@{}cl*{7}{c}@{\hspace{6pt}}*{7}{c}@{}}
      \toprule
      &
        & \multicolumn{7}{c}{\textbf{RE10K}}
        & \multicolumn{7}{c}{\textbf{DL3DV}} \\
      \cmidrule(lr){3-9}\cmidrule(lr){10-16}
        & & \multicolumn{3}{c}{Appearance} & \multicolumn{2}{c}{Camera} & \multicolumn{2}{c}{Depth}
        & \multicolumn{3}{c}{Appearance} & \multicolumn{2}{c}{Camera} & \multicolumn{2}{c}{Depth} \\
      \cmidrule(lr){3-5}\cmidrule(lr){6-7}\cmidrule(lr){8-9}
      \cmidrule(lr){10-12}\cmidrule(lr){13-14}\cmidrule(lr){15-16}
      $n$ & Gaussian head
        & PSNR$\uparrow$ & SSIM$\uparrow$ & LPIPS$\downarrow$ & AUC@$3^{\circ}\uparrow$ & AUC@$30^{\circ}\uparrow$ & $\delta_{1.25}\uparrow$ & AbsRel$\downarrow$
        & PSNR$\uparrow$ & SSIM$\uparrow$ & LPIPS$\downarrow$ & AUC@$3^{\circ}\uparrow$ & AUC@$30^{\circ}\uparrow$ & $\delta_{1.25}\uparrow$ & AbsRel$\downarrow$ \\
      \midrule
      \multirow{3}{*}{$2$} & Fully frozen & 22.14 & 0.754 & 0.233 & 26.4 & 78.8 & 0.833 & 0.161 & 22.01 & 0.685 & 0.267 & 30.6 & 88.8 & 0.790 & 0.179 \\
       & Image embedding only & 22.56 & 0.766 & 0.223 & 26.7 & 78.9 & 0.846 & 0.153 & 22.40 & 0.699 & 0.255 & 30.4 & 88.9 & 0.801 & 0.170 \\
       & Full finetuning & 22.88 & 0.775 & 0.216 & 26.8 & 78.8 & 0.854 & 0.148 & 22.69 & 0.708 & 0.248 & 30.2 & 88.8 & 0.807 & 0.166 \\
      \midrule
      \multirow{3}{*}{$3$} & Fully frozen & 24.07 & 0.806 & 0.190 & 25.6 & 78.8 & 0.846 & 0.150 & 24.05 & 0.747 & 0.212 & 34.1 & 88.8 & 0.803 & 0.163 \\
       & Image embedding only & 24.28 & 0.813 & 0.184 & 25.8 & 79.7 & 0.853 & 0.146 & 24.28 & 0.755 & 0.205 & 34.4 & 89.6 & 0.809 & 0.158 \\
       & Full finetuning & 24.47 & 0.818 & 0.180 & 26.0 & 80.0 & 0.857 & 0.143 & 24.50 & 0.761 & 0.200 & 34.6 & 90.0 & 0.814 & 0.155 \\
      \midrule
      \multirow{3}{*}{$5$} & Fully frozen & 25.24 & 0.834 & 0.165 & 25.0 & 78.8 & 0.849 & 0.143 & 25.21 & 0.779 & 0.184 & 33.6 & 87.5 & 0.814 & 0.155 \\
       & Image embedding only & 25.35 & 0.838 & 0.162 & 25.2 & 78.8 & 0.852 & 0.141 & 25.33 & 0.782 & 0.182 & 33.5 & 87.7 & 0.816 & 0.153 \\
       & Full finetuning & 25.47 & 0.842 & 0.159 & 25.3 & 78.8 & 0.855 & 0.139 & 25.44 & 0.785 & 0.180 & 33.4 & 87.5 & 0.818 & 0.151 \\
      \midrule
      \multirow{3}{*}{$7$} & Fully frozen & 25.86 & 0.850 & 0.151 & 24.7 & 78.8 & 0.852 & 0.138 & 25.68 & 0.795 & 0.170 & 33.9 & 86.3 & 0.816 & 0.152 \\
       & Image embedding only & 25.89 & 0.851 & 0.150 & 24.5 & 78.5 & 0.853 & 0.137 & 25.71 & 0.796 & 0.169 & 34.0 & 86.7 & 0.817 & 0.151 \\
       & Full finetuning & 25.96 & 0.853 & 0.148 & 24.5 & 78.8 & 0.855 & 0.135 & 25.76 & 0.798 & 0.168 & 34.1 & 87.5 & 0.818 & 0.150 \\
      \bottomrule
    \end{tabular}%
  }
  \endgroup
\end{table}

\subsection{Effect of Gaussian Head Finetuning}
\label{sec:app_gaussian_head_finetuning}

Table~\ref{tab:gaussian_head_finetuning} compares three Gaussian head configurations with AnySplat on RE10K and DL3DV. \emph{Fully frozen} keeps all pretrained Gaussian head weights fixed. \emph{Image embedding only} finetunes only a few image input layers in the Gaussian head, as described in Appendix~\ref{sec:app_appfusion}. \emph{Full finetuning} updates the entire Gaussian head using the same training hyperparameters as \emph{Image embedding only}.

Keeping the Gaussian head fully frozen yields performance close to image embedding finetuning on both datasets. The gap in appearance and depth quality becomes negligible as the number of observed views $n$ increases. We attribute this trend to the smaller spacing between input views along the same camera trajectory. Closer input views provide better coverage in the warped color prior, reducing the impact of holes and the need for appearance adaptation.

Finetuning the entire Gaussian head further improves appearance and depth quality, while camera accuracy remains broadly similar across the three configurations. These gains are modest, and both the frozen head and image embedding finetuning retain comparable performance.

%
%
%

\end{document}